%% file: main.tex
\documentclass[runningheads]{llncs_thmcountersep_modified}
\usepackage{algorithm,algpseudocode}
\algblockdefx{MRepeat}{EndRepeat}{\textbf{repeat}}{}
\algnotext{EndRepeat}
\algnotext{EndIf}
\algnewcommand{\algorithmicgoto}{\textbf{go to line}}%
\algnewcommand{\Goto}[1]{\algorithmicgoto~\ref{#1}}%

\algnotext{EndFor}
\algnotext{EndIf}
\algnotext{EndWhile}
\algnotext{EndFunction}
\algblockdefx{Do}{EndDo}{\textbf{do}\;}{}
\algnotext{EndDo}

\usepackage{marvosym}

\usepackage{graphicx} %
\usepackage{hyperref} %

\usepackage{amssymb}
\usepackage{amsmath}
\usepackage{comment}
    \includecomment{cav} %
    \excludecomment{arxiv} %
    \excludecomment{anon} %
    \includecomment{named} %

\usepackage{tabu}
\usepackage{multirow}
\usepackage{cite}
\usepackage{float}
\usepackage{wrapfig}
\usepackage{longtable}
\usepackage{paralist}
\usepackage{stmaryrd}
\usepackage{listings}
\usepackage[disable,colorinlistoftodos,textsize=footnotesize]{todonotes}

\newcommand\todoin[2][]{\todo[inline, caption={2do}, #1]{
\begin{minipage}{\textwidth-4pt}#2\end{minipage}}}

\usepackage{fancybox}

\usepackage{fontawesome}

\usepackage{tikz}
\usetikzlibrary{automata, positioning, arrows, shapes.geometric}
\tikzset{
->, %
>=stealth', %
node distance=2cm, %
every state/.style={thick}, %
initial text=$ $, %
}

\spnewtheorem{mytheorem}{Theorem}[section]{\bfseries}{\itshape} 
\spnewtheorem{mylemma}[mytheorem]{Lemma}{\bfseries}{\itshape}
\spnewtheorem{myproposition}[mytheorem]{Proposition}{\bfseries}{\itshape}
\spnewtheorem{mysublemma}[mytheorem]{Sublemma}{\bfseries}{\itshape}
\spnewtheorem{mycorollary}[mytheorem]{Corollary}{\bfseries}{\itshape}
\spnewtheorem{myfact}[mytheorem]{Fact}{\bfseries}{\itshape}
\spnewtheorem{mynotation}[mytheorem]{Notation}{\bfseries}{\rmfamily}
\spnewtheorem{myremark}[mytheorem]{Remark}{\bfseries}{\rmfamily}
\spnewtheorem{myexample}[mytheorem]{Example}{\bfseries}{\rmfamily}
\spnewtheorem{myassumption}[mytheorem]{Assumption}{\bfseries}{\rmfamily}
\spnewtheorem{mydefinition}[mytheorem]{Definition}{\bfseries}{\rmfamily}
\spnewtheorem{myrequirements}[mytheorem]{Requirements}{\bfseries}{\rmfamily}
\spnewtheorem{myproblem}[mytheorem]{Problem}{\bfseries}{\rmfamily}
\spnewtheorem*{myproof}{Proof}{\itshape}{\rmfamily}

\usepackage{apxproof}

\usepackage{array}
\newcolumntype{C}[1]{>{\centering\let\newline\\\arraybackslash\hspace{0pt}}m{#1}}
\newcolumntype{R}[1]{>{\raggedleft\let\newline\\\arraybackslash\hspace{0pt}}m{#1}}
\newcolumntype{L}[1]{>{\raggedright\let\newline\\\arraybackslash\hspace{0pt}}m{#1}}

\newcommand{\Av}{\mathrm{A\kern-0.09em v}}

\DeclareMathOperator*{\argmin}{arg\,min}

\newcommand{\bbN}{\mathbb{N}}

\newcommand{\bbR}{\mathbb{R}}

\newcommand{\bbZ}{\mathbb{Z}}

\newcommand{\calC}{\mathcal{C}}

\newcommand{\calH}{\mathcal{H}}

\newcommand{\calP}{\mathcal{P}}

\newcommand{\sfE}{\mathsf{E}}

\newcommand{\sol}{\mathsf{sol}}
\newcommand{\Vars}{\mathsf{Vars}}
\newcommand{\MyPr}{\mathsf{Pr}}
\newcommand{\cnt}{\mathsf{cnt}}
\newcommand{\thresh}{\mathsf{thresh}}
\newcommand{\rnd}{\mathsf{rnd}}

\newcommand{\Cell}[2]{\ensuremath{\mathsf{Cell}_{\langle #1, #2 \rangle}}}
\newcommand{\SatisfyingHashSet}[3]{\ensuremath{\mathsf{Cell}_{\langle #1, #2, #3 \rangle}}}
\newcommand{\Cnt}[2]{\ensuremath{\mathsf{Cnt}_{\langle #1, #2 \rangle}}}

\newcommand{\BoundedSAT}{\ensuremath{\mathsf{BoundedSAT}}}
\newcommand{\emptyList}{\ensuremath{\mathsf{emptyList}}}
\newcommand{\iter}{\ensuremath{\mathsf{iter}}}

\newcommand{\computeIter}{\ensuremath{\mathsf{computeIter}}}

\newcommand{\solCount}{\ensuremath{\mathsf{nSols}}}

\newcommand{\AddToList}{\ensuremath{\mathsf{AddToList}}}
\newcommand{\FindMedian}{\ensuremath{\mathsf{FindMedian}}}

\newcommand{\Lcnt}[1]{L_{#1}^{\mathsf{cnt}}}
\newcommand{\Ucnt}[1]{U_{#1}^{\mathsf{cnt}}}
\newcommand{\Lrnd}[1]{L_{#1}^{\mathsf{rnd}}}
\newcommand{\Urnd}[1]{U_{#1}^{\mathsf{rnd}}}
\newcommand{\Lout}[1]{L_{#1}^{\mathsf{out}}}
\newcommand{\Uout}[1]{U_{#1}^{\mathsf{out}}}

\newcommand{\boolDom}{\{0,1\}}
\usepackage{subcaption}

\usepackage[all,pdftex]{xy}
\SelectTips{cm}{}
\xyoption{rotate}

\begin{document}

\title{%
Systematic Parameter Decision in \\ Approximate Model Counting}
\titlerunning{Systematic Parameter Decision in Approximate Model Counting}

\begin{anon}%
    \author{
    Anonymous Authors
    }
    \authorrunning{Anonymous Authors
    }
    \institute{Omitted for Submission
    }
\end{anon}

\begin{named}%
    \author{
    Jinping Lei
    \and
    Toru Takisaka\textsuperscript{(\Letter)}
    \and
    Junqiang Peng
    \and
    Mingyu Xiao
    }
    \authorrunning{
    J.\ Lei et al.
    }
    \institute{
     University of Electronic Science and Technology of China, Chengdu, China \\ (\email{leijp@foxmail.com; takisaka@uestc.edu.cn; \\
    jqpeng0@foxmail.com; myxiao@uestc.edu.cn}) 
    }
\end{named}
\maketitle              %
\begin{abstract}
This paper proposes a novel approach to determining the internal parameters of the hashing-based approximate model counting algorithm $\mathsf{ApproxMC}$. In this problem, the chosen parameter values must ensure that $\mathsf{ApproxMC}$ is \emph{Probably Approximately Correct} (PAC), while also making it as efficient as possible. The existing approach to this problem relies on heuristics; in this paper, we solve this problem by formulating it as an optimization problem that arises from generalizing $\mathsf{ApproxMC}$’s correctness proof to arbitrary parameter values.

Our approach separates the concerns of algorithm soundness and optimality, allowing us to address the former without the need for repetitive case-by-case argumentation, while establishing a clear framework for the latter. Furthermore, after reduction, the resulting optimization problem takes on an exceptionally simple form, enabling the use of a basic search algorithm and providing insight into how parameter values affect algorithm performance. Experimental results demonstrate that our optimized parameters improve the runtime performance of the latest $\mathsf{ApproxMC}$ by a factor of 1.6 to 2.4, depending on the error tolerance.
\end{abstract}

\section{Introduction}\label{sect:intro}
The \emph{model counting problem}, 
also called $\mathsf{\#SAT}$, is the problem of computing the number of \emph{models} (or \emph{solutions}, \emph{satisfying assignments}) of a given propositional formula. 
Model counting is one of the most well-known $\mathsf{\#P}$-complete problems, 
where $\mathsf{\#P}$ is the class of counting problems whose decision counterparts are in $\mathsf{NP}$. 
Despite its computational hardness, the problem naturally arises in various application areas, such as control improvisation~\cite{GVF22}, network reliability~\cite{V79,DMPV17}, 
neural network verification~\cite{BSSM+19}, and probabilistic reasoning~\cite{R96,SBK05,CFMS+14,EGSS13a}.
Therefore, developing an efficient \emph{approximate} algorithm for model counting is of great interest to both theoreticians and practitioners.

One of the standard criteria for correctness in approximate algorithms is the \emph{Probably Approximately Correct} (PAC) criterion, also referred to as the $(\varepsilon, \delta)$-correct criterion.
In our context, a \emph{PAC approximate model counter} is a randomized algorithm that receives a triple $(F, \varepsilon,\delta)$, where $F$ is an input formula, $\varepsilon > 0$ is a \emph{tolerance parameter}, and $\delta \in (0,1]$ is a \emph{confidence parameter}, 
and returns a \emph{PAC approximation} of the model count of $F$. 
That is, with the probability at least $1-\delta$ (probably correct), the output should be within a factor of $(1 + \varepsilon)^{\pm 1}$ of the exact model count of $F$ (approximately correct).

\begin{wrapfigure}[8]{r}{0.35\textwidth}
\vspace{-1.8em}
\includegraphics[width=0.35\textwidth,clip]{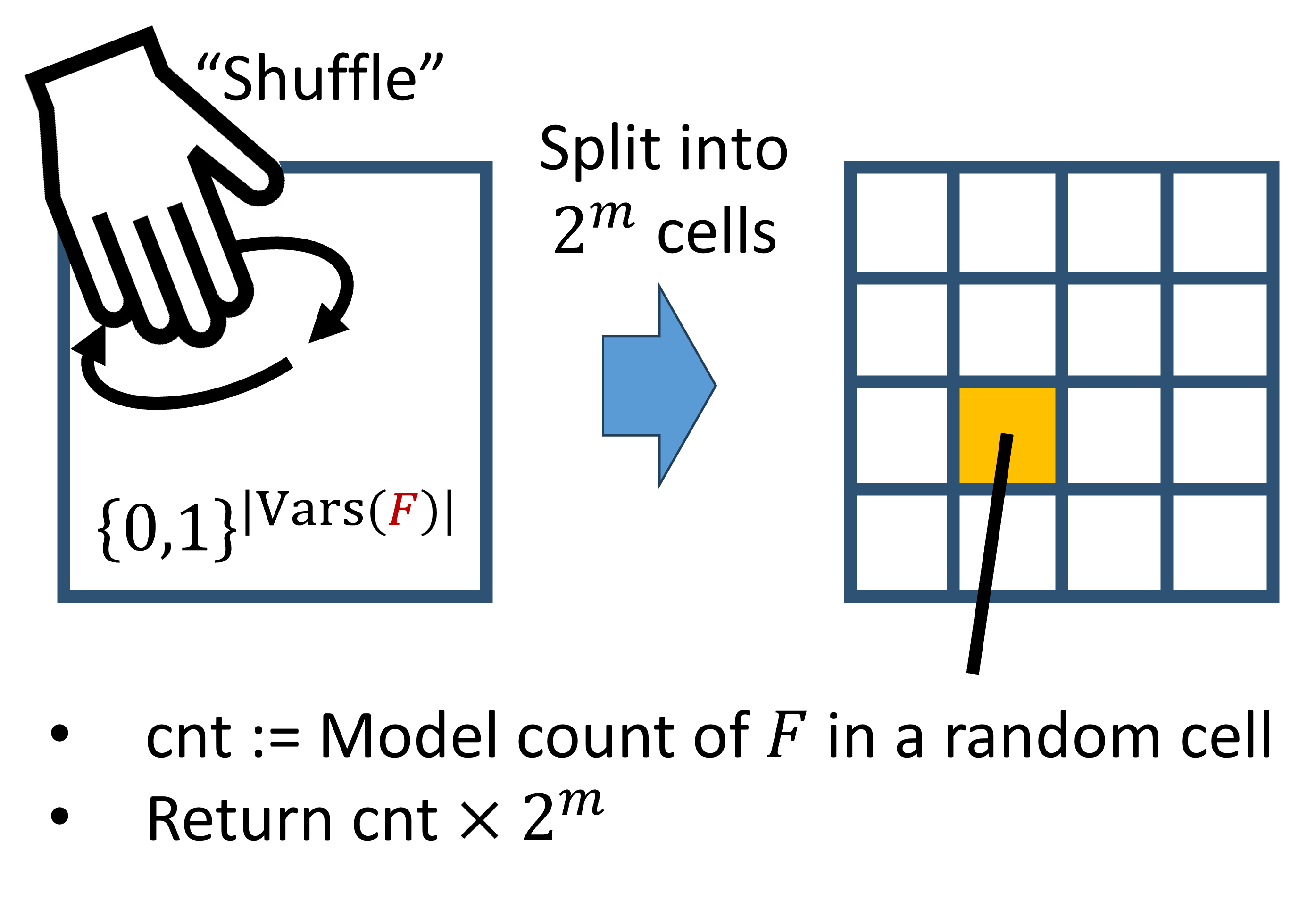}
\end{wrapfigure}
The current state-of-the-art in approximate model counting is a \emph{hashing-based} algorithm called $\mathsf{ApproxMC}$~\cite{ChakrabortyMV16}. 
Its high-level idea, which is also illustrated in the accompanying figure, is as follows: 
Given a formula $F$ with a set of variables $\Vars(F)$, 
we first randomly partition the model space 
$\{0,1\}^{|\Vars(F)|}$
into small subspaces (called \emph{cells}) using a hash function.  
We then compute the exact model count of $F$ within a randomly selected cell, and multiply it by the total number of cells to obtain an approximation of the model count of $F$. 
To efficiently achieve an $(\varepsilon,\delta)$-correct approximation %
for given $\varepsilon$ and $\delta$, 
$\mathsf{ApproxMC}$ iteratively runs a \emph{core counter} that performs the above estimation with low confidence (but shorter runtime) and returns the median of the collected estimates as its final result.

Since its initial proposal in 2013~\cite{ChakrabortyMV13}, 
researchers have shown sustained interest in improving the performance of hashing-based techniques~\cite{EGSS13b,EGSS13a,CFMS+14,IMMV16,MVCF+15,CMMV16,SM19,AM20,YM21,YCM22}.%
\todo{[MAYBE] would look nicer if we quickly summarize which kind of updates has been made here? To be updated if we have time. Also we may try to quickly explain why the problem persists for a decade despite its simple core structure? Toru understands it's because of the use of an NP oracle} 
The most recent update\footnote{
    We take $\mathsf{ApproxMC6}$~\cite{YangM23}  as our baseline, as it is the latest version that works for any $\varepsilon$ and $\delta$. 
    A newer version~\cite{PoteM025} exists, but it supports only $\varepsilon > 1$; see also \S\ref{sect:RelWorks}.
} 
to $\mathsf{ApproxMC}$~\cite{YangM23} significantly improved its efficiency with a novel \emph{rounding} technique, resulting in the sixth version,  $\mathsf{ApproxMC6}$.

\paragraph{A problem of concern: the parameter decision problem.} 
To determine the exact design of $\mathsf{ApproxMC}$, one must decide the values of its internal parameters, which may depend on the input values. 
For example, given an input $(F, \varepsilon, \delta)$, 
how many times should we run the core counter so that $\mathsf{ApproxMC}$ returns an $(\varepsilon,\delta)$-correct approximation? %
Also, the precise meaning of cells being 
``small'' must be specified by fixing another internal parameter $\thresh$. 
In addition to these, $\mathsf{ApproxMC6}$ requires yet another parameter $\rnd$, which is used in the rounding operation introduced in the recent update.
All these parameters---which interact in a rather delicate way---must be set so that $\mathsf{ApproxMC}$ is PAC; also, under this constraint, we wish $\mathsf{ApproxMC}$ to be as efficient as possible in its runtime. 
We call such a problem the  \emph{parameter decision problem} in $\mathsf{ApproxMC}$.

Existing works rely on heuristics to solve this problem. For example, it has been standard in $\mathsf{ApproxMC}$ to set $\thresh = 9.84(1+\frac{\varepsilon}{1+\varepsilon})(1+\frac{1}{\varepsilon})^2$ (plus one in some old versions). According to our analysis, this value was adopted not to optimize the runtime performance of $\mathsf{ApproxMC}$, but rather to simplify $\mathsf{ApproxMC}$'s correctness proof---it ensures that a certain intermediate parameter in the proof remains constant across different choices of $\varepsilon$. 

We are motivated to find a systematic approach to this problem for a couple of reasons.
First, recent additions of new features to $\mathsf{ApproxMC}$ have significantly increased the effort required to prove its correctness, and this increased effort could pose a potential threat to its long-term development.
With rounding added as a new feature, 
the key lemma for the correctness of $\mathsf{ApproxMC6}$~\cite[Lemma~4]{YangM23} now requires five different case analyses for its proof, %
demanding more than 8 pages to describe in LNCS format. %
This means that with every new version of $\mathsf{ApproxMC}$, we may have to revisit all these case analyses to ensure the correctness proof holds; 
or even worse, new features in the update may make the situation even more complicated. 

Second, the heuristic nature of the existing approach suggests that there might be a better parameter choice that improves the algorithm's efficiency. 
In fact, existing works do not provide a formal argument for the optimality of their chosen parameters; they have often been assumed without explanation as to why this specific value is used.

Third, we would like to make the relationship between the parameter choice and the resulting property of $\mathsf{ApproxMC}$ ``visible''. Specifically, we want to understand to what extent internal parameters can be modified while still maintaining $\mathsf{ApproxMC}$ to be PAC, and how such modifications affect the performance of the resulting algorithm. Understanding this inherent nature of $\mathsf{ApproxMC}$ will serve as the basis for its development. Unfortunately, existing studies offer quite limited information about this---they primarily focus on proving the correctness of $\mathsf{ApproxMC}$ under a specific parameter chosen in a top-down manner.

\paragraph{Contribution.} 
In this paper, 
we propose a systematic approach to the parameter decision problem
in %
$\mathsf{ApproxMC6}$. 
Our core idea is to recast the problem into an optimization problem. 
Specifically, we consider the following: 
\begin{inparaenum}[(a)]
\item A \emph{soundness condition}, which provides a sufficient condition on the internal parameters of $\mathsf{ApproxMC6}$ for the algorithm to be PAC, and
\item an objective function $\mathsf{Obj}$ over the set of all possible assignments to the internal parameters, which measures the runtime performance of $\mathsf{ApproxMC6}$ under a given assignment.
\end{inparaenum}
With these in place, we can find the desired parameter values by solving the optimization problem of minimizing 
$\mathsf{Obj}$   under the soundness condition.

There is a natural choice of $\mathsf{Obj}$ that serves as a proxy for the runtime performance of $\mathsf{ApproxMC6}$; therefore, 
the main technical challenge in our approach lies in identifying a soundness condition that admits a reasonable variety of parameter assignments.
We address this by extracting such a condition from the abstract structure of the existing proof of  $\mathsf{ApproxMC6}$'s correctness~\cite[Lemma~4]{YangM23}. 
At a high level, our approach is as follows: 

\medskip

\fbox{\parbox[c]{0.9\linewidth}{\it 
    Given a proof of the algorithm’s correctness under specific parameter values, we attempt to rewrite the proof while leaving the parameter values symbolic. This process naturally reveals the conditions under which the proof remains valid. We then use these as our soundness condition.
}}

\medskip

\noindent
In its original form, the resulting optimization problem is somewhat complex:  
its search space consists of eight-dimensional vectors with both real- and integer-valued entries,  
and its constraints (i.e., the soundness condition) are nonlinear.  
However, we show that the problem can be reduced to one with a much smaller two-dimensional, box-shaped search space (Corollary~\ref{cor:reducedOptProb}),  
over which even a brute-force search is feasible.  
Based on this reduction, we propose a new version of $\mathsf{ApproxMC6}$ that utilizes our optimal parameters, which we call $\mathsf{FlexMC}$.

\medskip
A crucial advantage of our approach is that it clearly separates the problem of ensuring soundness from that of achieving optimal runtime performance. 
This contrasts with conventional approaches, in which one typically first fixes a parameter that appears sound and reasonably near-optimal,  
and only then analyzes both its soundness and the resulting algorithm's efficiency.  
This separation makes our approach an effective and systematic solution to the parameter decision problem, as we elaborate below.
\begin{itemize}
    \item We settle the soundness problem by proving that $\mathsf{ApproxMC6}$ is PAC under any parameter assignment that satisfies the soundness condition (Theorem~\ref{thm:soundness}).
    This general approach eliminates the need for repetitive case analyses and significantly reduces the effort required.
    \item 
    By recasting the parameter decision problem as an optimization problem, we obtain a rigorous framework for identifying optimal parameters within the soundness condition, which turns out to be efficiently solvable via a simple search algorithm. Our experiments show that our optimal parameters make $\mathsf{ApproxMC6}$ from 1.65 to 2.46 times faster, depending on the value of $\varepsilon$.
    \item 
    Our reduced optimization problem visualizes the relationship between the parameter choices and the resulting runtime performance of $\mathsf{ApproxMC6}$, via our objective function $\mathsf{Obj}$ (Figure~\ref{fig:landscapes}). 
    It further reveals simple patterns in the behavior of some key intermediate parameters, such as the error probability bounds associated with the core counter, helping us interpret the behavior of $\mathsf{Obj}$ in light of how its input variables respond to parameter changes.
\end{itemize}

Another crucial advantage of our approach is its generality.
The parameter decision problem is fundamental in algorithm design, and our approach does not rely on any specific structure of $\mathsf{ApproxMC}$ or its correctness proof. As long as we have a proof---or even just a sketch---of correctness under specific parameter values, our approach lends itself to application. 
Because of this generality, our approach is likely to remain applicable to future revisions of $\mathsf{ApproxMC}$, and potentially even to the parameter decision problem in other algorithms.

\section{Preliminaries}\label{sect:preliminaries}
\paragraph{Model Counting, and PAC Model Counter.}
For a given propositional formula $F$ in Conjunctive Normal Form (CNF), we write $\Vars(F)$ to denote the set of variables that appear in $F$,  
and write $\sol(F)$ to denote the set of truth assignments to $\Vars(F)$ that make $F$ true (i.e., \emph{solutions} of $F$).

The \emph{model counting problem} is a problem to compute $|\sol(F)|$ for a given CNF formula $F$ as input. 
A \emph{Probably Approximately Correct} (PAC) model counter 
is a randomized algorithm $\mathsf{ApproxCount}(\cdot,\cdot,\cdot)$ that receives $(F,\varepsilon, \delta)$ as an input, where 
$F$ is a CNF formula, 
$\varepsilon > 0$ is a tolerance parameter, and
$\delta \in (0,1]$ is a confidence parameter; 
and returns an \emph{$(\varepsilon, \delta)$-approximation} of $|\mathsf{sol}(F)|$, i.e., a value $c$ that satisfies the following (when viewed as a random variable).
\begin{align}
    \MyPr\biggl[\frac{|\sol(F)|}{1+\varepsilon} \leq c \leq (1+\varepsilon)|\sol(F)|
    \biggr] 
    \geq 1-\delta \label{eq:EpsDelCorrectness}
\end{align}
We call the property (\ref{eq:EpsDelCorrectness}) the \emph{$(\varepsilon, \delta)$-correctness} of $\mathsf{ApproxCount}$.

    To simplify our argument, we do not present our theoretical results for \emph{projected model counting}, a generalized variant of 
    the model counting problem. 
    More concretely, for a given $F$ and a subset $\calP \subseteq \Vars(F)$, it is a problem to compute the number of solutions of $F$ projected on $\calP$. %
    Our argument is generalized to this problem in a canonical way, 
    and in fact, we treat this type of problems in our experiments.

\paragraph{Hash Families.} As the name of \emph{hashing-based} algorithm suggests, {\sf ApproxMC} uses hash functions as random seeds. 
For each $n,m \in \bbN$ with $m \leq n$, 
we assume access to a hash family $\calH(n,m)$ that consists of functions of the form $h: \boolDom^n \to \boolDom^m$. 
Since its second version~\cite{ChakrabortyMV16}, %
{\sf ApproxMC} exploits a relationship between $\calH(n,n)$ and $\calH(n,m)$ with $m <n$ that is characterized by \emph{prefix-slices}. 
Here, for a given $\Vec{x} = (x_1, \ldots, x_m, \ldots, x_n) \in \boolDom^n$ and $1 \leq m < n$, the \emph{$m^{\text{th}}$ prefix-slice} $\Vec{x}^{(m)}$ of $\Vec{x}$ is defined as $\Vec{x}^{(m)} = (x_1, \ldots, x_m) \in \boolDom^m$; 
and the \emph{$m^{\text{th}}$ prefix-slice} of a hash function $h:\boolDom^n \to \boolDom^n$ is defined as a function $h^{(m)}:\boolDom^n \to \boolDom^m$ such that $h^{(m)}(y) = h(y)^{(m)}$.
Then $\calH(n,m)$ is characterized as the set of all $m^{\text{th}}$ prefix-slices of $h\in \calH(n,n)$, 
i.e., $\calH(n,m) = \{h^{(m)} \mid h \in \calH(n,n)\}$. 
In this paper, we assume $\calH(n,m)$ with $m <n$ is always of this form.

Each \todo{[maybe] try to make the notation lighter}
hash function $h \in \calH(n,m)$ splits the set $\boolDom^n$ into $2^m$ disjoint subspaces, namely $\{h^{-1}(\alpha) \mid \alpha \in \boolDom^m\}$ (observe $h^{-1}(\alpha)$ can be empty). 
Such subspaces are called \emph{cells}.
For a given CNF formula $F$ with $|\Vars(F)| =n$, a hash function $h:\boolDom^n \to \boolDom^m$ and $\alpha \in \boolDom^m$, the set of solutions of $F$ that belong to the cell $h^{-1}(\alpha)$ is denoted by $\SatisfyingHashSet{F}{h}{\alpha}$; that is, we let $\SatisfyingHashSet{F}{h}{\alpha} := \sol(F) \cap h^{-1}(\alpha)$. 
We are mostly interested in the cardinality of %
$\SatisfyingHashSet{F}{h}{\alpha}$ for randomly chosen $h$ and $\alpha$, which are 
generated from $h' \in \calH(n,n)$ and $\alpha' \in \boolDom^n$ by taking their $m^{\text{th}}$ prefix-slices; 
so we mostly omit them from the notation and write $\Cell{F}{m}$. 
We also write $\Cnt{F}{m}$ to denote the value $|\Cell{F}{m}|$. 

Observe, for a given $h \in \calH(n,n)$, the sequence of its prefix-slices $h^{(1)}, \ldots, h^{(n)}$
can be seen as an iterative splitting policy of the assignment space $\boolDom^n$: 
That is, $\boolDom^n$ is divided into $(h^{(1)})^{-1}(0)$ and $(h^{(1)})^{-1}(1)$; then $(h^{(1)})^{-1}(x)$ is divided into $(h^{(2)})^{-1}(x,0)$ and $(h^{(2)})^{-1}(x,1)$ for each $x \in \{0,1\}$, and so on. 
Then each $\alpha \in \boolDom^n$ naturally induces a sequence of cells 
$(h^{(1)})^{-1}(\alpha^{(1)})\supseteq \ldots \supseteq (h^{(n)})^{-1}(\alpha^{(n)})$, 
which can be seen as a sequence of decisions on which cell to take under the iterative splitting policy $h$. 
In particular, for any fixed $h \in \calH(n,n)$, $\alpha \in \boolDom^n$ and $1\leq m' < m \leq n$, we have $\Cell{F}{m} \subseteq \Cell{F}{m'}$, and thus $\Cnt{F}{m} \leq \Cnt{F}{m'}$.

For its $(\varepsilon, \delta)$-correctness, 
{\sf ApproxMC} demands the following property to its hash families: 
For given $n\in \bbN$, a formula $F$ with $|\Vars(F)| = n$, and $m \in \{1, \ldots, n\}$, we demand the following holds, where expectation and variance are considered with respect to the uniform distribution over $\calH(n,m)$:%
\begin{align}\label{eq:hashFamilyCoreProp}
    \sfE[\Cnt{F}{m}]  = \frac{|\sol(F)|}{2^m} 
    \quad , \quad 
    \sigma^2[\Cnt{F}{m}] \leq \sfE[\Cnt{F}{m}].
\end{align}
We assume our hash families satisfy (\ref{eq:hashFamilyCoreProp}). 
There is a standard realization of hash families that satisfy all the properties above, called the \emph{XOR hash families}\cite{gomes2006near}.
In our theoretical analysis, they can be anything that satisfy these properties. %

\begin{algorithm}[t]
	\caption{{\sf ApproxMC6}$(F, \varepsilon, \delta)$}
      \label{alg:ApproxMC6}
	\begin{algorithmic}[1]
        \State $(\thresh, \rnd, p_L, p_U) \leftarrow \mathsf{SetParameters}(\varepsilon,\delta);$ \quad  $t \leftarrow \mathsf{ComputeIter}(p_U, p_L, \delta)$; \label{ln:initBegins}
		\State $Y \leftarrow \BoundedSAT(F, \thresh);$ 
		\If{$(|Y| < \thresh)$} \Return $|Y|;$ \EndIf
		\State $C\leftarrow\emptyList; \iter \leftarrow 0;$ \label{ln:initEnds}
		\Repeat 
		\State $\solCount \leftarrow {\sf ApproxMC6Core}(F, \thresh, \rnd);$
		\State $\AddToList(C,\solCount);$ \ $\iter \leftarrow \iter + 1;$
		\Until{$(\iter \ge t)$}$;$ 
		\State finalEstimate $\leftarrow \FindMedian(C);$
		\State \Return finalEstimate$;$		
	\end{algorithmic}
\end{algorithm}
\begin{algorithm}[t]
	\caption{{\sf ApproxMC6Core}$(F,\thresh,\rnd)$}
	\label{alg:ApproxMC6Core}
	
	\begin{algorithmic}[1]
		\State Choose $h$ at random from $\calH(n,n);$ \label{ln: roundmc hash}
		\State Choose $\alpha$ at random from $\{0,1\}^n;$ \label{ln: roundmc hash value} 
        \State $\Cnt{F}{n} \leftarrow \BoundedSAT\left(F\wedge \left(h^{(n)}\right)^{-1}\left(\alpha^{(n)}\right), \thresh\right);$\label{line:sanityCheck}
        \If{$\Cnt{F}{n} \geq \thresh$} $m \leftarrow n;$ \Return $2^{n};$ \EndIf \label{line:singularOutput}
		\State $m \leftarrow \mathsf{LogSATSearch}(F, h, \alpha, \thresh);$ \label{line:LogSATSearch}
		\State $\Cnt{F}{m} \leftarrow \BoundedSAT\left(F\wedge \left(h^{(m)}\right)^{-1}\left(\alpha^{(m)}\right), \thresh\right);$ \label{ln: roundmc count cell}
		\State \Return $(2^m\times \max\{\Cnt{F}{m},\rnd\});$ \label{line:coreCounterOutput}
	\end{algorithmic}
\end{algorithm}

\paragraph{The {\sf ApproxMC} Algorithm.}
A pseudocode of {\sf ApproxMC6}~\cite{YangM23}, the latest version of {\sf ApproxMC}, is given in Algorithm~\ref{alg:ApproxMC6} (some details are modified from the original description for a better explanation). 
Its high-level idea is as follows: \todo{[maybe] guess its better to mention SetParameters somewhere in this paragraph?}
It iteratively calls the core counter $\mathsf{ApproxMC6Core}$ for precomputed times, namely $t$ times. 
For each invocation, $\mathsf{ApproxMC6Core}$ returns an approximate model count of the input formula $F$ which is $\varepsilon$-accurate with the confidence lower than $1-\delta$. 
Therefore, by taking the median of sufficient number of independent outputs from the core counter, $\mathsf{ApproxMC6}$ obtains an $(\varepsilon,\delta)$-correct estimate of $\sol(F)$. 
To compute the iteration number $t$, 
we first compute upper bounds $p_L$ and $p_U$ on the probability that $\mathsf{ApproxMC6Core}$ \emph{under-} or \emph{over-estimates} $|\sol(F)|$, respectively (i.e., it returns a value less than $\frac{1}{1+\varepsilon}|\sol(F)|$, or more than $(1+\varepsilon)|\sol(F)|$). 
These values are determined by an intricate theoretical argument, which is the main analysis target of this paper; a more detailed discussion will be done in~\S\ref{sect:overview}. 
Then $t$ is computed by the $\mathsf{computeIter}$ procedure, 
which computes the smallest odd number that is sufficient to guarantee $(\varepsilon,\delta)$-correctness of Algorithm~\ref{alg:ApproxMC6}, given $p_L$ and $p_U$ as probability bounds. See Appendix~\ref{append:prelim} for its formal definition. %

The core counter $\mathsf{ApproxMC6Core}$ works as follows\footnote{
The original algorithm in~\cite{YangM23} always round up the value of $\Cnt{F}{m}$ in Line~\ref{line:coreCounterOutput} when $\varepsilon \geq 3$. We find this arrangement is not necessary in our argument, so we omit it. \label{footnote:alwaysRound}
}. 
It first samples a hash function $h$ and a %
string $\alpha$ randomly. 
After a sanity check (Line~\ref{line:sanityCheck}-\ref{line:singularOutput}), 
it finds the smallest number $m$ such that 
$\Cnt{F}{m} < \thresh$,\todo{[maybe]try to better signpost what cnt and cell were} meaning that $\Cell{F}{m}$ is 
``small'' %
(Line~\ref{line:LogSATSearch}). 
Then it computes $\Cnt{F}{m}$, 
possibly ``round'' it by $\rnd$, and return it after multiplying it by $2^m$.
Here, %
$\mathsf{LogSATSearch}$~\cite{ChakrabortyMV16}  
efficiently finds the number $m$ via a combination of linear and galloping searches. 
The subroutine $\mathsf{BoundedSAT}$ employs a SAT solver to enumerate distinct solutions of $F$ until either it enumerates all solutions, 
or the number of enumerated solutions reaches $\thresh$. 
Notice that $\mathsf{BoundedSAT}$ makes up to $\thresh$ times of SAT calls, and thus, the number of calls to $\mathsf{BoundedSAT}$ is a determining factor of the performance of $\mathsf{ApproxMC6}$. 
The \emph{rounding} operation (Line~\ref{line:coreCounterOutput}) is the new feature of the latest update~\cite{YangM23}; 
it lets us to derive a tighter probability bounds $p_L$ and $p_U$, and thus offers a smaller repetition number $t$. 
Apart from that, the high-level structure of $\mathsf{ApproxMC6Core}$ remains the same as the initial version of the core counter.

\section{Parameter Decision Problem as Optimization}\label{sect:overview}
In this section, we give a technical overview of how to recast the 
parameter decision problem into an optimization problem.
We begin by stating the problem for %
$\mathsf{ApproxMC6}$,  in a form that is still slightly informal.

\medskip

\fbox{\parbox[c]{0.9\linewidth}{
    {\bf Key Problem (Parameter decision problem):} 
    For a given pair $(\varepsilon, \delta)$, how should we choose the values of $(\thresh, \rnd, p_L, p_U)$—that is, the output of $\mathsf{SetParameters}(\varepsilon,\delta)$—so that Algorithm~\ref{alg:ApproxMC6} becomes an $(\varepsilon, \delta)$-correct model counter and is as efficient as possible?
}}

\medskip

The conventional approach~\cite[Lemma~4]{YangM23} is to first fix $\thresh$ and $\rnd$ heuristically, 
and then derive error probability bounds $p_L$ and $p_U$ using what we call the \emph{bounding argument}. 
The core of our approach is to generalize this argument into a form that can accommodate arbitrary parameter assignments; 
as explained in~\S\ref{sect:intro}, this consideration naturally leads to our \emph{soundness condition}, i.e., a sufficient condition on the internal parameters that ensures $\mathsf{ApproxMC6}$ is PAC. Once we obtain such a condition, we can search for the most favorable sound parameter assignment via optimization, guided by any preferred objective function.
For a fixed $\varepsilon > 0$, the bounding argument proceeds as follows. %
\begin{enumerate}
        \item Take any sound parameter assignment $\vec{c}$. The parameters consist of $\thresh$, $\rnd$, and those used in the internal argument to derive $p_L$ and $p_U$. 
        For a CNF formula $F$, let $L(F)$ and $U(F)$ denote the events that the output of  $\mathsf{ApproxMC6Core}(F, \thresh, \rnd)$ under- and over-estimates $|\sol(F)|$, respectively (we will formally define what we mean by \emph{event} in \S\ref{sect:formalizeIntoOptimization}). 
    \item Claim $L(F) \subseteq L'(\vec{c},F)$ and $U(F) \subseteq U'(\vec{c},F)$ for any $F$. Here, $L'(\vec{c},F)$ and $U'(\vec{c},F)$ are particular events whose probability of occurrence can be evaluated by standard concentration inequalities (Lemma~\ref{lem:concentration} in Appendix~\ref{append:formalizeIntoOptimization}). Correctness of the claim is derived from the soundness condition; inspired from how we prove it, we call this claim the \emph{cut-off argument}. %
    \label{item:CutoffArgument}
    \item Claim $\MyPr[L'(\vec{c},F)] \leq p_L(\vec{c})$ and $\MyPr[U'(\vec{c},F)] \leq p_U(\vec{c})$ for any $F$, where $p_L(\vec{c})$ and $p_U(\vec{c})$ are the values that naturally arise by applying concentration inequalities on $\MyPr[L'(\vec{c},F)]$ and $\MyPr[U'(\vec{c},F)]$, respectively. 
    The correctness of the inequalities is guaranteed by that of the concentration inequalities.
\end{enumerate}

Notice that, when we determine the exact details of the bounding argument, we begin from Step~\ref{item:CutoffArgument}:
that is, we first specify the description of $L'(\vec{c},F)$ and $U'(\vec{c},F)$ for arbitrary $\vec{c}$ and $F$ according to the existing proof, and then figure out our soundness condition by considering under which assignment $\vec{c}$ the 
cut-off argument holds. 
Below, we give an overview of how $L'(\vec{c},F)$ is constructed; the construction of $U'(\vec{c},F)$ is similar at the high-level, while there are some differences in details. 
There, we fix $\varepsilon > 0, F$, and $\vec{c}$ (which includes $\thresh$ and $\rnd$); let $n = |\Vars(F)|$; and write $L$ and $L'$ instead of $L(F)$ and $L'(\vec{c},F)$, respectively.

\begin{itemize}
    \item 
    Let $\Lout{i}$ be 
        \todo{[maybe] consider cutting the superscript out. I put this only to avoid the unmatching notation use in CAV23}
    the event $L$ conditioned that 
    $\mathsf{ApproxMC6Core}(F,\thresh,\rnd)$ terminates with $m=i$, for $i \in \{1,\ldots,n\}$.
    Then we have
    $L = \Lout{1} \cup \cdots \cup \Lout{n}$.
    \item %
    We then ``cut $\Lout{i}$ off'' from $L$ for each $i$ that is either too small or too large: that is, we claim the following for some $m^\downarrow$ and $m^\uparrow$ such that $m^\downarrow < m^\uparrow$. 
    \[
    \Lout{1} \cup \cdots \cup \Lout{m^\downarrow} \subseteq T_{m^\downarrow} 
    \quad , \quad 
    \Lout{m^\uparrow} \cup \cdots \cup \Lout{n} = \emptyset.
    \]
    Here, $T_i$ is the event that $\Cnt{F}{i} < \thresh$ holds (i.e., cells are small enough if we let $m = i$). 
    Feasible choices for such $m^\downarrow$ and $m^\uparrow$ depend on the value of $|\sol(F)|$. 
    A key observation here %
    is that 
    we can describe these values by a single reference number $m^*$ that depends on $|\sol(F)|$: that is, we let 
     $m^\downarrow := m^* -k^\downarrow$ and $m^\uparrow := m^* -k^\uparrow$, where $k^\downarrow$ and $k^\uparrow$ are fixed parameters. 
    Roughly speaking, $m^*$ estimates the number such that $2^{m^*} \times \thresh$ approximates $|\sol(F)|$ the best. %
    Now, we claim the event $L$ is subsumed by the event
    \begin{align}
        T_{m^\downarrow} \cup \Lout{m^\downarrow+1} \cup \cdots \cup \Lout{m^\uparrow-1}. \label{eq:OverApproxL}
    \end{align}
    \item
    To apply concentration inequalities
    (cf. Lemma~\ref{lem:concentration}, Appendix~\ref{append:formalizeIntoOptimization}), we perform additional reductions on each clause in (\ref{eq:OverApproxL}), reducing the entire event (\ref{eq:OverApproxL}) into $L'$. 
    Finally, we claim %
    $L \subseteq L'$; see~(\ref{eq:cutoffArgumentL}) for its exact form.

\end{itemize}
We observe four parameters appear in the argument above, namely $\thresh$, $\rnd$, $k^\downarrow$, and $k^\uparrow$ ($\rnd$ implicitly appears in %
$L$); 
and the exact definition of $m^*$ involves another parameter $a$, which we call the \emph{shifting parameter}. 
In addition, we also do a similar %
argument to fix the description of  $U'(\vec{c},F)$, which demands another copy of %
$k^\downarrow$, $k^\uparrow$, $a$ (they may take different values from the ones for $L$).  
Overall, 
we have $\Vec{c} = (\thresh, \rnd, a_L$, $k_L^\downarrow, k_L^\uparrow, a_U, k_U^\downarrow, k_U^\uparrow)$ 
as our internal parameters. %

By investigating the structure of the existing proof~\cite[Lemma~4]{YangM23}---in particular, 
by observing how it proves the cut-off argument %
under their particular parameter value---we determine our soundness condition as the conjunction of $\varphi_1,\ldots,\varphi_4$ given in Definition~\ref{def:soundness}. 
With any objective function $\mathsf{Obj}$ in mind, now we recast the parameter decision problem as the following optimization problem:
\begin{align}
    \mbox{\bf Minimize} 
    \quad \mathsf{Obj}(p_L(\Vec{c}), p_U(\Vec{c}), \Vec{c}) 
    \qquad
    \mbox{\bf Subject to} \quad \varphi_1(\Vec{c}) \land \ldots \land \varphi_4(\Vec{c}).\label{ourAbstOptimizationProb}
\end{align} 
This can be understood as the parameter decision problem in the following way. For a given $(\varepsilon,\delta)$, 
one can seek for a solution $\Vec{c}_{\text{sol}}$ to (\ref{ourAbstOptimizationProb}) w.r.t. $\varepsilon$; 
once it is found, then 
we return $(\thresh,\rnd,p_L(\Vec{c}_{\text{sol}}),p_U(\Vec{c}_{\text{sol}}))$, where $\thresh$ and $\rnd$ are taken from $\Vec{c}_{\text{sol}}$. 
It makes Algorithm~\ref{alg:ApproxMC6} $(\varepsilon,\delta)$-correct, by definition; and the resulting algorithm is the most efficient w.r.t. $\mathsf{Obj}$, and within the scope of the bounding argument.

In this paper, we adopt 
$\mathsf{Obj}_1^{(\delta)} (\Vec{c}) = 
\mathsf{ComputeIter}(p_L(\Vec{c}), p_U(\Vec{c}),\delta) \times \thresh$ 
as the default choice of $\mathsf{Obj}$, where $\delta$ should be taken from the input to $\mathsf{ApproxMC6}$. 
This appears to be one of the most natural choices to evaluate the runtime performance of Algorithm~\ref{alg:ApproxMC6}, if not unique, 
as it makes $\mathcal O(t\times \thresh\times \log_{2}|\Vars(F)|)$-times SAT calls during its run \cite{ChakrabortyMV16}. 

\section{Technical Details and Soundness}\label{sect:formalizeIntoOptimization}
In this section, we give the technical details of how we formalize the bounding argument. 
In particular, 
we give an exact description of 
our cut-off argument (relations~(\ref{eq:cutoffArgumentL}) and~(\ref{eq:cutoffArgumentU})), 
probability bounds $p_L$ and $p_U$ (equations~(\ref{eq:DefOfProbBounds})), 
and the soundness conditions (Definition~\ref{def:soundness}). 
Based on these, 
we prove the soundness of $p_L$ and $p_U$ as probability bounds (Theorem~\ref{thm:soundness}), 
meaning that $p_L(\Vec{c})$ and $p_U(\Vec{c})$ indeed over-approximate 
the probabilities of events $L$ and $U$ %
whenever $\Vec{c}$ satisfies the soundness conditions. 
By these results, we have an exact description of our optimization problem (Definition~\ref{def:ourOptimizationProb}), as well as its soundness (Corollary~\ref{cor:soundnessOptProb}).

\paragraph{Events.} 
In what follows, 
 we give various statements that involve events in the $\mathsf{ApproxMC6Core}$ (cf. an overview on the bounding argument in~\S\ref{sect:formalizeIntoOptimization}). 
 Here we formally define them. Recall, for an input $(F, \thresh, \rnd)$ with $ |\Vars(F)| = n$,  $\mathsf{ApproxMC6Core}$ randomly samples a hash function $h\in \calH(n,n)$ and a vector $\alpha \in \boolDom^n$; 
therefore, we identify an event with the set of all tuples $(h, \alpha) \in \calH(n,n) \times \boolDom^n$ 
that trigger the event %
in  $\mathsf{ApproxMC6Core}$. 
The probability $\MyPr[E]$ of an event $E$ is then canonically understood as 
$\MyPr[E] = \frac{|E|}{|\calH(n,n)| \times 2^n}$.

Now we define our events under $(F, \thresh, \rnd)$ and $\varepsilon$, or simply events, as follows. 
For a lighter notation, we suppress the use of $F, \thresh, \rnd$ and $\varepsilon$ in the notation. 
When we need to clarify them, 
we say e.g. ``$\MyPr[L] \leq p_L(\Vec{c})$ holds under $(F, \thresh, \rnd)$ and $\varepsilon$''. 
Often we only need to clarify $F$, in which case, we say ``$\MyPr[L] \leq p_L(\Vec{c})$ holds under $F$''.
We also write $\cnt_m$ and $\sol$ to denote $\Cnt{F}{m}$ and $\sol(F)$, respectively. 
Finally, we let $n = |\Vars(F)|$ and $i \in \{1, \ldots, n\}$ below.

\begin{enumerate}
    \item An event $T_i$ refers to the event  $\cnt_i < \thresh$; for convenience, we let $T_0$ be an empty event. 
    \item Events $\Lcnt{i}$ and $\Ucnt{i}$ refer to 
    the events where the value $\cnt_i \times 2^i$ under- or over-estimates $|\sol|$, respectively; that is, 
     $\Lcnt{i}$ means 
    $\cnt_i \times 2^i < \frac{1}{1+\varepsilon} |\sol|$, and 
    $\Ucnt{i}$ means $\cnt_i \times 2^i > (1+\varepsilon) |\sol|$. 
    \item Events $\Lrnd{i}$ and $\Urnd{i}$ refer to the events where the value $\rnd \times 2^i$ under- or over-estimates $|\sol|$, respectively; that is, 
    $\Lrnd{i}$ means 
    $\rnd \times 2^i < \frac{1}{1+\varepsilon} |\sol|$, and $\Urnd{i}$ means $\rnd \times 2^i > (1+\varepsilon) |\sol|$. 
    \item An event $\Lout{i}$ refers to 
    the event where the output of $\mathsf{ApproxMC6Core}$ under-estimates $|\sol|$ with $m = i$; that is,  
    $\Lout{i} = T_i \cap \overline{T_{i-1}} \cap \Lcnt{i} \cap \Lrnd{i}$. 
    Similarly, its over-estimation variant $\Uout{i}$ is defined by 
    $\Uout{i} = T_i \cap \overline{T_{i-1}} \cap (\Ucnt{i} \cup \Urnd{i})$.
    \item An event $\hat{U}$ refers to the event where the singular output $2^n$ by the core counter over-estimates $|\sol|$ (cf. Line~\ref{line:singularOutput} in Algorithm~\ref{alg:ApproxMC6Core}); that is, $\hat{U}$ means $2^n > (1+\varepsilon)|\sol|$ (we do not consider the $L$-variant because it never happens).
    \item Finally, $L$ and $U$ refer to the events where the output of $\mathsf{ApproxMC6Core}$  under- or over-estimates $|\sol|$, respectively. 
    They are also written as $L = \Lout{1} \cup \cdots \cup \Lout{n}$ and $U = \Uout{1} \cup \cdots \cup \Uout{n} \cup (\overline{T_n} \cap \hat{U})$, respectively.
\end{enumerate}

\paragraph{Some Definitions.}
We write $\bbR_{\geq a}$ and $\bbR_{>a}$ to denote the sets $[a,\infty)$ and $(a, \infty)$, respectively. 
We fix the types of parameters\footnote{We use these specific lower bounds on $\thresh$ and $\rnd$ to get a simpler representation of some theorems we prove. There is no merit to weaken these bounds. We also do not miss any optimal parameters if we require $\thresh$ to be a natural number; we use reals to match the argument with existing works.}  by 
$\thresh \in \bbR_{\geq 2}, \rnd\in \bbR_{\geq 1}, a_L, a_U \in \bbR_{>0}$, and $k_L^\downarrow, k_L^\uparrow, k_U^\downarrow, k_U^\uparrow \in \bbZ$. 
Thus, our parameter space is 
$\calC = \bbR_{\geq 2} \times\bbR_{ \geq 1} \times\bbR_{>0} \times\bbZ \times\bbZ \times\bbR_{>0} \times\bbZ \times\bbZ$, 
whose elements are of the form $\Vec{c} = (\thresh, \rnd, a_L, k_L^\downarrow$, $ k_L^\uparrow$, $a_U, k_U^\downarrow, k_U^\uparrow)$. 
As an obvious requirement to make the cut-off arguments well-defined, we assume 
$k_L^\downarrow > k_L^\uparrow \land k_U^\downarrow > k_U^\uparrow$. 
In what follows, we assume $\varepsilon >0$ and $\Vec{c} \in \calC$ are given unless specified, and parameters in our presentation (e.g., $\thresh, \rnd,\ldots$) refer to the values in $\Vec{c}$. 
We do not fix a formula $F$ meanwhile; this is because we would like to argue $p_L(\Vec{c})$ and $p_U(\Vec{c})$ are upper bounds of $\MyPr[L]$ and $\MyPr[U]$, respectively, under \emph{any} $F$.

On events under $F$ with $|\Vars(F)| = n$, 
we let $\Lcnt{i},\Ucnt{i},\Lrnd{i}$, $\Urnd{i}$ be all empty for $i \not\in \{1, \ldots, n\}$; 
we also let $T_i$ be empty for $i <1$, and be the full set  $\calH(n,n) \times \boolDom^n$ for $i >n$. 
This is for technical convenience; it makes the definition of cut-off arguments well-defined even if some cut-off points do not belong to $\{1, \ldots, n\}$ (see~(\ref{eq:cutoffArgumentL}) and~(\ref{eq:cutoffArgumentU})). 

Finally, we define a crucial notion in the cut-off argument, namely the \emph{reference number} $m^*$. 
For a given CNF formula $F$ with $|\Vars(F)| =n$ and 
 $Q \in \{L, U\}$, we define $m_Q^*$ as the smallest $m \in \bbZ$ that satisfies $2^{-m} \times |\sol(F)| < a_Q \times \thresh$. Similar to events, we suppress %
 $F, \thresh$, etc. in its notation.

\paragraph{Cut-off Arguments.} 
We define our cut-off arguments as the following relationships. 
We note again that, for a given $F$, 
their validity under $(F, \thresh, \rnd)$ and $\varepsilon$ depends on the choice of $\Vec{c}$ and $\varepsilon$. 
\begin{align}
    L &\subseteq T_{m_L^*-k_L^\downarrow} 
    \cup
    \Lcnt{m_L^*-(k_L^\downarrow-1)} 
    \cup \cdots \cup 
    \Lcnt{m_L^*-(k_L^\uparrow+1)}, \label{eq:cutoffArgumentL} \\
    U &\subseteq 
    \Ucnt{m_U^*-(k_U^\downarrow-1)} \cup \cdots \cup \Ucnt{m_U^*-(k_U^\uparrow+1)} \cup (\overline{T}_{m_U^*-k_U^\uparrow} \cup \Ucnt{m_U^*-k_U^\uparrow}). \label{eq:cutoffArgumentU}
\end{align}

\paragraph{Probability Bounds $p_L$ and $p_U$.} 
By applying standard concentration inequalities (Lemma~\ref{lem:concentration} in Appendix~\ref{append:formalizeIntoOptimization}), we can bound the probability of each clause in the RHS of~(\ref{eq:cutoffArgumentL}) and~(\ref{eq:cutoffArgumentU}), as follows. 
There, $q_T$ and $q_{\overline{T}\cup U}$ only provide a trivial bound $1$ in a certain case; 
this corresponds to the situation where Lemma~\ref{lem:concentration} is not applicable. %
A proof is given in Appendix~\ref{appendLem:ConcreteConcentration}.
\begin{mylemma}\label{lem:ConcreteConcentration}
    For given $\thresh >0$ and $\varepsilon >0$, let
    \begin{align*}
    q_T(a,k) &= 
        \begin{cases}
        \frac{1}{1+ (1-\frac{1}{(a \times 2^{k-1})})^2 \times a \times 2^{k-1} \times \thresh}        & \text{if } a \times 2^{k-1} > 1,\\  
        1    & \text{otherwise},
    \end{cases}\\
    q_L(a,k) &= \frac{1}{1+ (1-\frac{1}{(1+\varepsilon)})^2 \times a \times 2^{k-1} \times \thresh}, \\
    q_U(a,k) &= \frac{1}{1+ \varepsilon^2 \times a \times 2^{k-1} \times \thresh}, \\
    q_{\overline{T}\cup U}(a,k) &= 
        \begin{cases}
            \max\bigl\{
                \frac{1}{1+ (\frac{1}{(a \times 2^{k})}-1)^2 \times a \times 2^{k-1} \times \thresh}, q_U(a,k)
            \bigr\}        & \text{if } a \times 2^{k} < 1,\\  
            1    & \text{otherwise}.
        \end{cases}\\
    \end{align*}
    Then for any formula $F$,  $k\in \bbZ$, $Q \in \{L, U\}$, and $a_Q >0$ (which in turn specify $m_Q^*$), we have the following under $F,$ $\thresh$, and $\varepsilon$ (and independent of $\rnd$):
    \begin{align*}
        \MyPr[T_{m_Q^* - k}] &\leq q_T(a_Q, k),& \MyPr[\Lcnt{m_Q^* - k}] &\leq q_L(a_Q, k), \\
        \MyPr[\Ucnt{m_Q^* - k}] &\leq q_U(a_Q, k),& \MyPr[\overline{T_{m_Q^* - k}} \cup \Ucnt{m_Q^* - k}] &\leq q_{\overline{T}\cup U}(a_Q, k). \tag*{\qed}
    \end{align*}
\end{mylemma}

Based on the argument (\ref{eq:cutoffArgumentL}), (\ref{eq:cutoffArgumentU}) and Lemma~\ref{lem:ConcreteConcentration}, we have the concrete description of upper bounds $p_L(\Vec{c})$ and $p_U(\Vec{c})$ as follows:
\begin{align}
\begin{split}
    p_L(\Vec{c}) &= q_T(a_L, k_L^\downarrow) + q_L(a_L, k_L^\downarrow -1) + \ldots + q_L(a_L, k_L^\uparrow+1), \\
    p_U(\Vec{c}) &= q_U(a_U, k_U^\downarrow-1) + \ldots + q_U(a_U, k_U^\uparrow+1) + q_{\overline{T}\cup U}(a_U, k_U^\uparrow). \label{eq:DefOfProbBounds}
\end{split}
\end{align}
Here, if $k_L^\downarrow = k_L^\uparrow +1$ then we read $p_L(\Vec{c}) = q_T(a_L, k_L^\downarrow)$; if $k_U^\downarrow -1= k_U^\uparrow$ then we read $p_U(\Vec{c}) = q_{\overline{T}\cup U}(a_U, k_U^\uparrow)$. 
Observe, if (\ref{eq:cutoffArgumentL}) and (\ref{eq:cutoffArgumentU}) hold under a given formula $F$, then we have $\MyPr[L] \leq p_L(\Vec{c})$ and $\MyPr[U] \leq p_U(\Vec{c})$ under that $F$.

\paragraph{Soundness Conditions.}
Now let us clarify when the cut-off arguments (\ref{eq:cutoffArgumentL}) and (\ref{eq:cutoffArgumentU}) hold. 
Showing conclusion first, our soundness conditions are the following.
\begin{mydefinition}[soundness conditions]\label{def:soundness}
We call the following $\varphi_1, \ldots, \varphi_4$ the \emph{soundness conditions} for the cut-off arguments:
    \begin{align*}
    \varphi_1(\Vec{c}) &\equiv k_L^\downarrow > k_L^\uparrow \land k_U^\downarrow > k_U^\uparrow, &
    \varphi_2(\Vec{c}) &\equiv  \rnd \geq \frac{a_L}{1+\varepsilon}\times 2^{k_L^\uparrow} \times \thresh, \\
    \varphi_3(\Vec{c}) &\equiv \rnd \leq (1+\varepsilon)a_U \times  2^{k_U^\uparrow -1} \times \thresh, &
    \varphi_4(\Vec{c}) &\equiv \frac{1}{a_U}\times 2^{-(k_U^\downarrow -1)} \leq 1+\varepsilon. %
\end{align*}
\end{mydefinition}

We write $\varphi_i^{(\varepsilon)}$ instead of $\varphi_i$ when we need to make the underlying $\varepsilon$ explicit. 
The condition $\varphi_1$ is the obvious one we mentioned in the beginning. 
The conditions $\varphi_2,\varphi_3$, and $\varphi_4$ %
derive the following %
under any $F$ with $|\Vars(F)| = n$.
\begin{enumerate}
    \item \label{item:phi2}
    $\varphi_2(\Vec{c})$ cuts off the upper clauses of $L = \Lout{1}\cup \cdots \cup \Lout{n}$. %
    That is, it implies
    $\Lout{i} = \emptyset$  for each  $i \geq m_L^*-k_L^\uparrow$. 
    \item \label{item:phi3}
    $\varphi_3(\Vec{c})$ eliminates the rounding clause $\Urnd{i}$ from $\Uout{i} = T_i \cap \overline{T_{i-1}} \cap (\Ucnt{i} \cup \Urnd{i})$.  That is, it implies
    $\Urnd{i} = \emptyset$ for each $i \leq m_U^*-k_U^\uparrow$.
    \item \label{item:phi4}
    $\varphi_4(\Vec{c})$ cuts off the lower clauses of $U = \Uout{1} \cup \cdots \cup \Uout{n} \cup (\overline{T_n} \cap \hat{U})$. More concretely, 
    $\varphi_4(\Vec{c})$ implies $T_i \cap \Ucnt{i} = \emptyset$ for each $i \leq m_U^*-k_U^\downarrow$ and thus, together with $\varphi_1(\Vec{c})$ and $\varphi_3(\Vec{c})$, it implies 
    $\Uout{i} = \emptyset$ for each $i \leq m_U^*-k_U^\downarrow$.
\end{enumerate}

For a given $\varepsilon > 0$, we say $\Vec{c}$ is \emph{$\varepsilon$-valid}, or simply valid, if $\varphi_i^{(\varepsilon)}(\Vec{c})$ holds for each $i\in\{1, \ldots, 4\}$; 
or $\Vec{c}$ is \emph{($\varepsilon$-)invalid} otherwise. Now we have the following; the proof is given in Appendix~\ref{appendThm:soundness}. 
\begin{mytheorem}[soundness of $p_L(\Vec{c})$ and $p_U(\Vec{c})$ as probability bounds]\label{thm:soundness}
    For a given $\varepsilon > 0$, 
    suppose $\Vec{c} = (\thresh,\rnd,a_L, k_L^\downarrow, k_L^\uparrow, a_U, k_U^\downarrow, k_U^\uparrow)$ is an $\varepsilon$-valid vector. 
    Then for an arbitrary CNF formula $F$, 
    the relationships (\ref{eq:cutoffArgumentL}) and (\ref{eq:cutoffArgumentU}) hold under $(F, \thresh, \rnd)$ and $\varepsilon$, and thus,  
    we have 
    $\MyPr[L] \leq p_L(\Vec{c})$ and $\MyPr[U] \leq p_U(\Vec{c})$ under $(F, \thresh, \rnd)$ and $\varepsilon$, 
    where $p_L(\Vec{c})$ and $p_U(\Vec{c})$ are as defined in (\ref{eq:DefOfProbBounds}). \qed
\end{mytheorem}

We conclude this section with the exact description of our optimization problem. 
At this point, our objective function $\mathsf{Obj}$ can be any function over $\bbR_{\geq 0} ^2 \times \calC$.  
\begin{mydefinition}[parameter decision problem for {\sf ApproxMC6}]\label{def:ourOptimizationProb}
For $\varepsilon>0$, 
the \emph{parameter decision problem for {\sf ApproxMC6}} w.r.t. $\mathsf{Obj}$ is defined as follows:
    \begin{align}
    \underset{\Vec{c} \in \calC}{\mbox{\bf Minimize}}
    \quad \mathsf{Obj}(p_L(\Vec{c}), p_U(\Vec{c}), \Vec{c}) 
    \qquad
    \mbox{\bf Subject to} \quad %
    \bigwedge_{1\leq i \leq 4} \varphi_i^{(\varepsilon)}(\Vec{c}).\label{line:ourOptimizationProb}
\end{align}
\end{mydefinition}

By Theorem~\ref{thm:soundness}, any feasible $\Vec{c}$ in~(\ref{line:ourOptimizationProb}) gives us sound bounds of failure probabilities of $\mathsf{ApproxMC6Core}$ (under the matching $\varepsilon$). Formally, we have the following.
\begin{mycorollary}[soundness of Problem~(\ref{line:ourOptimizationProb})]\label{cor:soundnessOptProb}
    For a given $\varepsilon >0$, 
    if a vector $\Vec{c} = (\thresh,\rnd,a_L, k_L^\downarrow, k_L^\uparrow, a_U, k_U^\downarrow, k_U^\uparrow) \in \calC$ is feasible in Problem~(\ref{line:ourOptimizationProb}), 
    then for an arbitrary CNF formula $F$, we have 
    $\MyPr[L] \leq p_L(\Vec{c})$ and $\MyPr[U] \leq p_U(\Vec{c})$ under $(F, \thresh, \rnd)$ and $\varepsilon$, 
    where $p_L(\Vec{c})$ and $p_U(\Vec{c})$ are as defined in (\ref{eq:DefOfProbBounds}). \qed
\end{mycorollary}

\section{Finding The Optimum in Two Steps}\label{sect:problemReduction}
So far we have clarified the exact description of our optimization problem in (\ref{line:ourOptimizationProb}).
The problem looks rather complex at a glance; 
in this section, however, 
we prove we can reduce it into a significantly simpler form under rather mild assumption on $\mathsf{Obj}$ (Corollary~\ref{cor:reducedOptProb}). 
The reduced problem is a box-constrained optimization problem over two-dimensional vectors $(\thresh, a_U)$, and  
together with the non-triviality condition on $p_U$ (cf. Lemma~\ref{lem:ConcreteConcentration}), 
the range of $a_U$ becomes $\frac{1}{1+\varepsilon} \leq a_U <1$. 
This makes the search space small enough to find a global optimum even via the brute-force search, or if preferred, more sophisticated search algorithms.
In addition, the assumption on $\mathsf{Obj}$ seem to be satisfied by any reasonable $\mathsf{Obj}$, see Assumption~\ref{ourAssumption}. 

We propose a simple search algorithm to find an optimum of Problem (\ref{line:ourOptimizationProb}) under our default objective function $\mathsf{Obj}_1^{(\delta)} (\Vec{c}) = 
\mathsf{ComputeIter}(p_L(\Vec{c}), p_U(\Vec{c}),\delta) \times \thresh$ (Algorithm~\ref{alg:FindOptPara}). 
In addition, Our reduction result %
make the parameter dependency \todo{[maybe] the word "parameter dependency" not so good? here we talk about the relationship between parameters and prob bounds} ``visible''; 
it reduces the dimension of our search space from eight to two, revealing rather simple behavior of the probability bounds over the reduced search space, as demonstrated in Figure~\ref{fig:landscapes}.

\subsection{Problem Reduction}
Our results in this section require the following assumption on  $\mathsf{Obj}$.

\begin{myassumption}\label{ourAssumption}
    The objective function $\mathsf{Obj}$ satisfies the following: 
    \begin{inparaenum}[(a)]
    \item It is non-decreasing with respect to $p_L(\Vec{c})$ and $p_U(\Vec{c})$, and
    \item it does not directly refer to the values of $\Vec{c}$ except for $\thresh$, i.e., it is of the form $\mathsf{Obj}(p_L(\Vec{c}), p_U(\Vec{c}), \thresh)$. 
\end{inparaenum}
\end{myassumption}
These assumptions are rather mild; in fact, it seems any reasonable $\mathsf{Obj}$ should satisfy them. 
It is counterintuitive that $\mathsf{Obj}$ violates the first condition, as it means using looser probability bounds can be beneficial to improve the performance of $\mathsf{ApproxMC}$.  
The second condition also looks natural to require, as all parameters in $\Vec{c}$ other than $\thresh$ can affect the number of calls to an $\mathsf{NP}$ oracle only by affecting the value of $p_L(\Vec{c})$ and $p_U(\Vec{c})$, which in turn affects the repetition number $t$ of $\mathsf{ApproxMC6Core}$. 
Meanwhile, $\thresh$ can directly affect it because it is fed to $\mathsf{BoundedSAT}$; thus it is natural to keep it accessible for $\mathsf{Obj}$. 

\medskip
Our reduction is done by shrinking the search space $\calC$ into a smaller space, 
that is, 
we show we only need to consider vectors that satisfy certain properties, say $\Psi$. 
This is done by showing that, for any valid $\Vec{c} \in \calC$, we can always find 
another valid vector $\vec{d}$ that satisfies $\Psi$ and also realizes a smaller value of $\mathsf{Obj}$. 
Here, Assumption~\ref{ourAssumption} 
enables us to reason about the second property of $\vec{d}$ without looking into the concrete definition of $\mathsf{Obj}$; 
it suffices to check
\begin{inparaenum}[(a)]
\item $\Vec{d}$ is valid, 
\item $p_L(\Vec{c}) \geq p_L(\Vec{d})$ and $p_U(\Vec{c}) \geq p_U(\Vec{d})$, and
\item $\Vec{c}$ and $\Vec{d}$ have the same value of $\thresh$. 
\end{inparaenum}
The result %
is given as follows; the proof is given in Appendix~\ref{appendThm:NarrowingDownTheSearchSpace}.

\begin{mytheorem}[search space shrinking]\label{thm:NarrowingDownTheSearchSpace}
    Suppose $\sf{Obj}$ satisfies Assumption~\ref{ourAssumption}.
    For a given $\varepsilon>0$, consider the following constraints on $\Vec{c} \in \calC$: 
    \begin{align*}
    \psi_1(\Vec{c}) &\equiv k_L^\uparrow = 0 \land k_U^\downarrow = 1 \land  k_U^\uparrow = 0, &
    \psi_2(\Vec{c}) &\equiv k_L^\downarrow = 1  \lor k_L^\downarrow = 2, \\
    \psi_3(\Vec{c}) &\equiv a_L = \frac{(1+\varepsilon)^2}{2}a_U, &
    \psi_4(\Vec{c}) &\equiv \rnd = \frac{(1+\varepsilon)a_U}{2} \times \thresh.
\end{align*}
Then there exists a solution $\Vec{c}_{\text{sol}}$ to the optimization problem (\ref{line:ourOptimizationProb}) with the same $\varepsilon$ that satisfies $\psi_1(\Vec{c}_{\text{sol}})\land \psi_2(\Vec{c}_{\text{sol}})\land \psi_3(\Vec{c}_{\text{sol}})\land \psi_4(\Vec{c}_{\text{sol}})$. \qed
\end{mytheorem}

Observe $\psi_1 \land \psi_2 \land \psi_3 \land \psi_4$ implies $\varphi_1 \land \varphi_2 \land \varphi_3$, while reducing $\varphi_4$
to $a_U \geq \frac{1}{1+\varepsilon}$. 
Thus we have our reduced problem as follows; 
see Appendix~\ref{append:reducedOptProb} for a 
proof.

\begin{mycorollary}[a reduced problem of~(\ref{line:ourOptimizationProb})]\label{cor:reducedOptProb}
    For given $\thresh$, $a_U$ and $k_L^\downarrow$, define \todo{[maybe]Probably we can use concrete description or $\hat{p}_L$ etc. at this point, it may make the presentation simpler}
    $\Vec{c}(\thresh,a_U,k_L^\downarrow)$ by
    \[
    \Vec{c}(\thresh,a_U,k_L^\downarrow) = 
    (\thresh, \frac{(1+\varepsilon)a_U}{2} \times \thresh,\frac{(1+\varepsilon)^2}{2}a_U,k_L^\downarrow,0,a_U,1,0),
    \]
    and let $\hat{p}_Q(\thresh, a_U) = \min\{p_Q(\Vec{c}(\thresh,a_U,1)),p_Q(\Vec{c}(\thresh,a_U,2))\}$, where $Q\in\{L,U\}$. Then for $\mathsf{Obj}$ that satisfies  Assumption~\ref{ourAssumption}, the global minimum of the following optimization problem coincides with that of (\ref{line:ourOptimizationProb}):
\begin{align}
\begin{split}
    \underset{(\thresh, a_U)}{\mbox{\bf Minimize}}
    &\quad \mathsf{Obj}(\hat{p}_L(\thresh, a_U), \hat{p}_U(\thresh, a_U), \thresh) \\
    \mbox{\bf Subject to}
    &\quad
    \thresh \geq 2, 
     \quad 
    a_U \geq \frac{1}{1+\varepsilon}. 
    \label{ourReducedOptimizationProb}
\end{split}
\end{align}
Also, if $(\thresh, a_U)$ is a solution to~(\ref{ourReducedOptimizationProb}), and  
$k \in \{1,2\}$ satisfies $\hat{p}_L(\thresh, a_U) = p_L(\Vec{c}(\thresh,a_U,k))$, 
then $\Vec{c}(\thresh,a_U,k)$ is a solution to~(\ref{line:ourOptimizationProb}). \qed
\end{mycorollary}

\subsection{The Landscape of Probability Bounds and Objective Function}\label{subsect:landscape}
\begin{figure*}[!tb]
\centering
\begin{subfigure}[b]{0.48\textwidth}
\includegraphics[width=\textwidth]{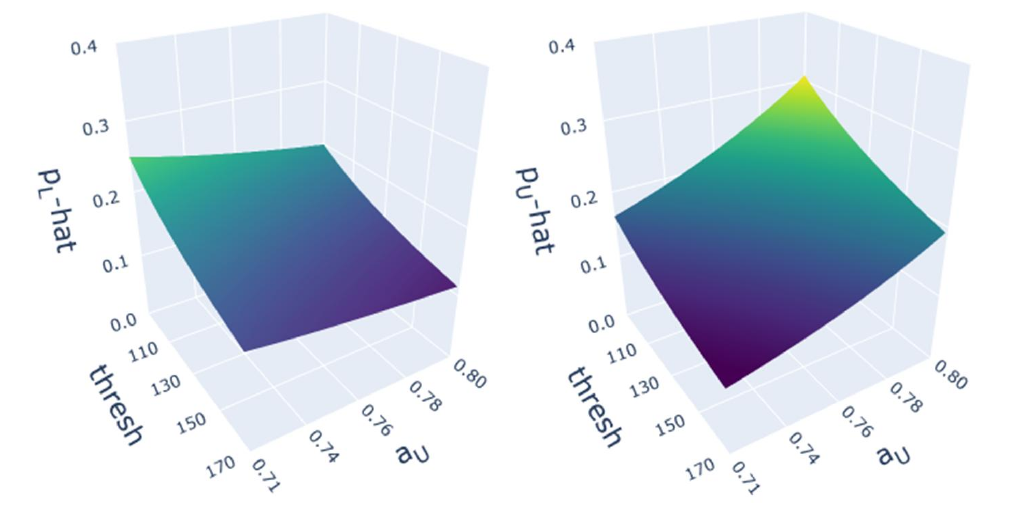}
\caption{Error probability bounds $\hat{p}_L$, $\hat{p}_U$}
\label{fig:errorBoundLandscape}
\end{subfigure}~~~~~~%
\begin{subfigure}[b]{0.48\textwidth}
\includegraphics[width=\textwidth]{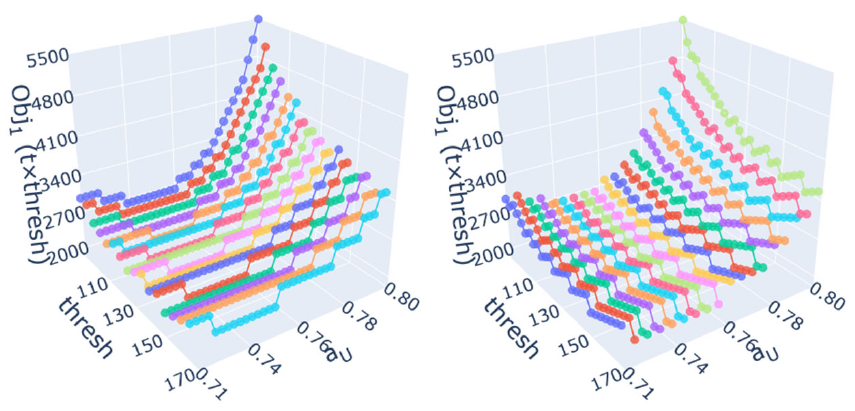}
\caption{Default objective function $\mathsf{Obj}_1$}
\label{fig:errorBoundSimulation}
\end{subfigure}~~~~~~%
\caption{Behavior of $\hat{p}_L$, $\hat{p}_U$, and $\mathsf{Obj}_1$ under $(\varepsilon,\delta) = (0.4,0.001)$. 
The left and right in Figure~\ref{fig:errorBoundSimulation} show the slices of $\mathsf{Obj}_1$ along fixed $\thresh$'s and $a_U$'s, respectively. 
}
\label{fig:landscapes}
\end{figure*}

Before introducing our search algorithm, %
we visualize the behavior of our probability bounds and default objective function over the reduced search space in %
Figure~\ref{fig:errorBoundLandscape} and~\ref{fig:errorBoundSimulation}, respectively. 
To be precise, 
our reduced search space is 
$\{(\thresh, a_U) \mid \thresh \geq 2,  a_U \geq \frac{1}{1+\varepsilon} )\}$; and 
Figure~\ref{fig:errorBoundLandscape} shows the behavior of $\hat{p}_L$ and $\hat{p}_U$, and Figure~\ref{fig:errorBoundSimulation} shows that of the value 
$\mathsf{Obj}_1^{(\delta)}(\hat{p}_L(\thresh, a_U)$, $\hat{p}_U(\thresh, a_U), \thresh)$. 
For brevity, we simply write the latter $\mathsf{Obj}_1(\thresh, a_U)$. %

The figures only show snapshots that highlight the parameters' global 
behavior.
We note the case $a_U \geq 1$ is out of our interest because it implies $\hat{p}_U=1$; 
also we can canonically bound $\thresh$ from above in the search algorithm (\S\ref{subsect:solutionSearch}). 

\medskip

Figure~\ref{fig:errorBoundLandscape} visualizes simple relationships between the probability bounds and input parameters, which is theoretically anticipated: $\hat{p}_L$ and $\hat{p}_U$ 
are decreasing and increasing, respectively, w.r.t. $a_U$; 
we also observe $\hat{p}_L$ and $\hat{p}_U$ are both decreasing w.r.t. $\thresh$. 
This behavior suggests us not much further reduction is possible ``for free'', as we are now in the realm of trade-off (observe a larger value of $\thresh$ induces more SAT calls by $\BoundedSAT$, and thus it will act negatively on any reasonable $\mathsf{Obj}$). 
Therefore, to find a solution to~(\ref{ourReducedOptimizationProb}), we 
 fix a concrete $\mathsf{Obj}$ and resort to search algorithms. \todo{[maybe]try to mention the balancing ability}

While the behavior of $\mathsf{Obj}_1$ is more complex, it is still a combination of simple one-dimensional behaviors, as shown in Figure~\ref{fig:errorBoundSimulation}. For a fixed $\thresh'$, the function $\mathsf{Obj}_1(\thresh', \underline{\quad})$ behaves in a unimodal manner; for a fixed $a_U'$, the behavior of $\mathsf{Obj}_1(\underline{\quad}, a_U')$ looks like a sawblade, as $\mathsf{Obj}_1$ is of the form $t \times \thresh$, where $t$ decreases in a discrete manner as $\thresh$ increases.
As discussed in the end of \S\ref{sect:overview}, $\mathsf{Obj}_1$ serves as a proxy for the runtime performance of $\mathsf{ApproxMC6}$; therefore, one can estimate from Figure~\ref{fig:errorBoundSimulation} how different choices of the parameters affect the resulting performance of $\mathsf{ApproxMC6}$.

\subsection{Comparison with Conventional Probability Bounds}
\begin{figure}[t]
\begin{minipage}{0.37\columnwidth}
    \centering
    \vspace{2em}
    \includegraphics[width=0.9\textwidth]{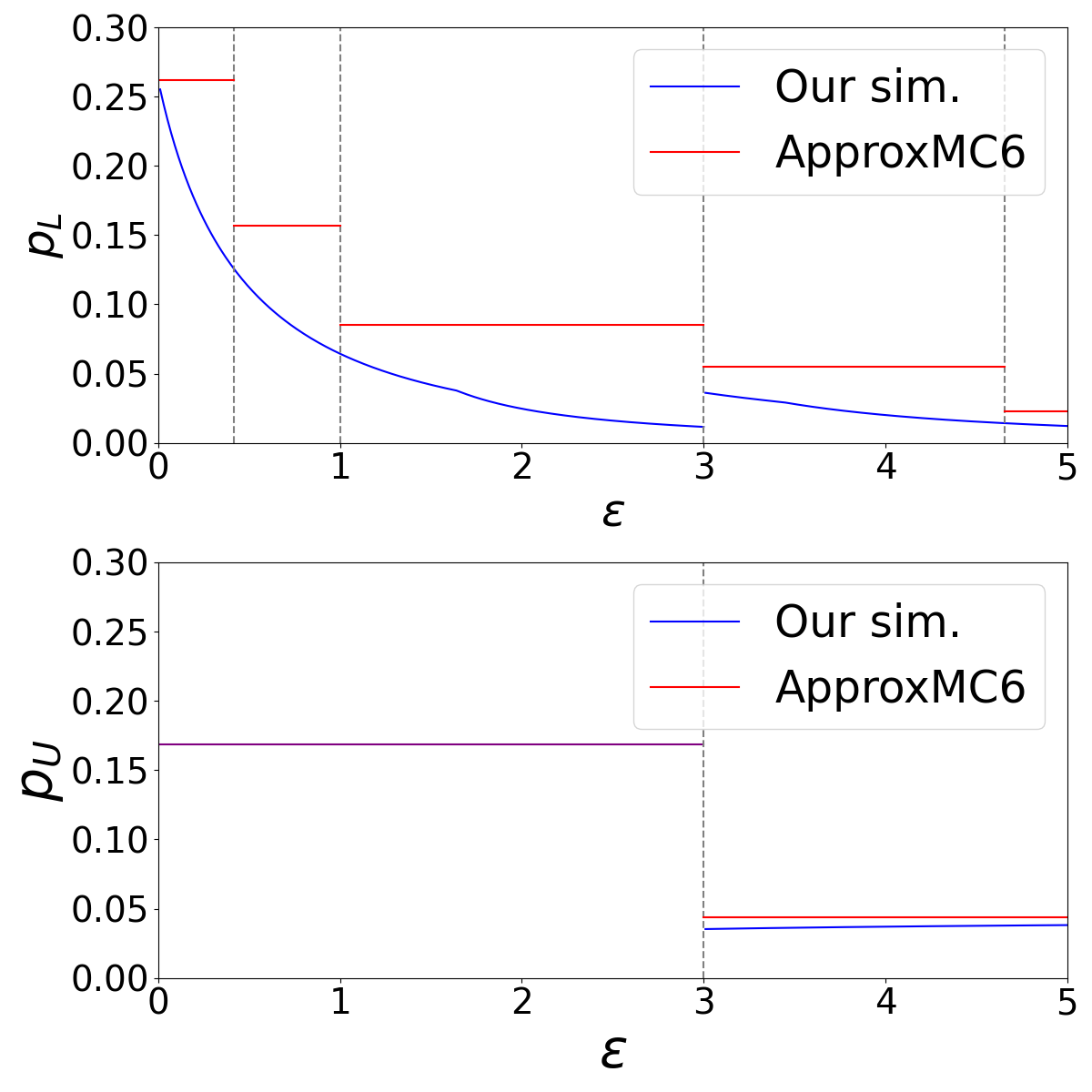}
    \caption{Bound comparison}
    \label{fig:unimodality}
  \end{minipage}%
\hspace{0.04\columnwidth}
\begin{minipage}{0.54\columnwidth}
    \begin{algorithm}[H]
	\caption{{\sf findOptParams}$(\varepsilon, \delta)$}
      \label{alg:FindOptPara}
	\begin{algorithmic}[1]
        \State $\thresh^* \gets 2$;
        \State $(t^*, a_U^*)\gets${\sf FindOptIter}$(\thresh^*, \varepsilon, \delta)$;
		\State $\thresh\gets 2$;
		\Repeat 
		\State $(t, a_U)\gets$ {\sf FindOptIter}$(\thresh, \varepsilon, \delta)$;
        \If{$t^*\times \thresh^*> t\times \thresh$}
            \State $\thresh^*\gets \thresh$; $t^*\gets t$; $a_U^*\gets a_U$;
        \EndIf
		\State $\thresh\gets \thresh + 1$;
		\Until{$(\thresh \ge \thresh^*\times t^*)$}$;$ 
        \State \Return $(\thresh^*, t^*, a_U^*)$;\label{line:setParamReturnedVal}
	\end{algorithmic}
\end{algorithm}
  \end{minipage}%
\end{figure}

Figure~\ref{fig:unimodality}
compares the probability bounds used in $\mathsf{ApproxMC6}$~\cite{YangM23} and our bounds under specific $\thresh$ and $a_U$ referenced from~\cite{YangM23}, so that the latter simulates the former to some extent. %
More specifically, our bounds in Figure~\ref{fig:unimodality} are generated by letting 
$\thresh = 9.84(1+\frac{\varepsilon}{1+\varepsilon})(1+\frac{1}{\varepsilon})^2$, 
and 
$a_U = (1+\frac{\varepsilon}{1+\varepsilon})^{-1}$ if $\varepsilon \leq 3$,
or 
$a_U = \frac{1}{2} \times (1+\frac{\varepsilon}{1+\varepsilon})^{-1}$ if $\varepsilon > 3$\footnote{
    Our parameters $a_L$ and $a_U$ are related to the one in~\cite{YangM23} called $\mathsf{pivot}$, via an equation $\mathsf{pivot} = a_Q \times \mathsf{thresh} = 9.84(1+\frac{1}{\varepsilon})^2$ for $Q \in \{L,U\}$ (\cite{YangM23} uses the same $\mathsf{pivot}$ to evaluate $L$ and $U$). 
    This is why we let $a_U = (1+\frac{\varepsilon}{1+\varepsilon})^{-1}$. Also, multiplying $a_U$ with $\frac{1}{2}$ is equivalent to decrementing $k_U^\downarrow$ and $k_U^\uparrow$, see the proof of Theorem~\ref{thm:NarrowingDownTheSearchSpace}. 
}, in $\hat{p}_L$ and $\hat{p}_U$. 
Our bounds uniformly outperform the conventional one; 
furthermore, having Theorem~\ref{thm:soundness}, 
our bounds %
are shown to be sound by merely observing $\thresh \geq 2$ and $a_U \geq \frac{1}{1+\varepsilon}$. 
This is in contrast to the soundness proof in~\cite{YangM23} that demands fine-tuned case analysis.  
The key point in our argument is to give the maximum freedom on the 
parameter values 
at the beginning. 
By this, it eventually turns out that $a_U$ can take care of all the dependencies between other parameters.

\subsection{Solution Search}\label{subsect:solutionSearch}
Now we present a simple algorithm to search for a solution to~(\ref{ourReducedOptimizationProb}) 
under $\mathsf{Obj}_1$.
Our search algorithm is given in Algorithm~\ref{alg:FindOptPara}. 
Its high-level idea is that we enumerate the optimal value of $t$ for each fixed $\thresh$. 
We search such a $t$ for a fixed $\thresh$ 
by performing 
a search algorithm (brute-force, or a more sophisticated algorithm if preferred) over 
$a_U \in [\frac{1}{1+\varepsilon}, 1]$ 
with the objective function 
$\computeIter(\hat{p}_L(a_U,\thresh),\hat{p}_U(a_U,\thresh),\delta)$. 
Recall it is safe to assume $\thresh$ is a natural number;
also, by $t \geq 1$, we can dynamically bound the range of $\thresh$ by the current best value of $\mathsf{Obj}_1$. Wrapping these up, we have our search algorithm as Algorithm~\ref{alg:FindOptPara}. 

\paragraph{Answering Key Problem.} 
Finally, we come back to our Key Problem: 
For given $(\varepsilon,\delta)$, what should we return as an output of $\mathsf{SetParameters}(\varepsilon,\delta)$? 
Now we answer: We solve Problem~(\ref{ourReducedOptimizationProb}) by 
Algorithm~\ref{alg:FindOptPara} and get $(\thresh^*, t^*, a_U^*)$, recover $\rnd$ by letting 
$\rnd^* = \frac{(1+\varepsilon)a_U^*}{2} \times \thresh^*$ (cf. $\psi_4$ in Theorem~\ref{thm:NarrowingDownTheSearchSpace}), 
and return $(\thresh^*, \rnd^*, \hat{p}_L(\thresh^*, a_U^*), \hat{p}_U(\thresh^*, a_U^*))$. 
By Corollary~\ref{cor:reducedOptProb}, 
these values are characteristics of a solution to the original parameter decision problem~(\ref{line:ourOptimizationProb}), and thus, the resulting $\mathsf{ApproxMC6}$ is $(\varepsilon,\delta)$-correct by Theorem~\ref{thm:soundness}, and has the minimum $t \times \thresh$ within the scope of the bounding argument. We call Algorithm~\ref{alg:ApproxMC6} $\mathsf{FlexMC}$ when it features this novel $\mathsf{SetParameters}$ procedure.

\input{Sec_Experiments}

\input{Sec_Related_Works}

\section{Conclusion}
In this paper, we proposed a systematic approach for the parameter decision problem in the correctness proof of approximate model counting algorithms. 
In our approach, the clear separation of soundness and optimality problems enables us to
solve the former in one-shot, while also offering a rigorous approach to the latter. 
Additionally, the reduced form of the parameter decision problem clarifies the simplicity of parameter dependency and the behavior of error probability bounds for $\mathsf{ApproxMCCore}$, making the correctness proof's inherent nature more evident.

\medskip
\noindent
{\bf Acknowledgment.} 
We thank Mate Soos and Jiong Yang for their help in finding the necessary resources for our experiments.

\bibliographystyle{splncs04}
\bibliography{ref}

\newpage

\appendix

\section*{Appendix}
\todoin{
Todo's toward submission
\begin{itemize}
    \item Terminologies:
    \begin{itemize}
    \item Fix the capitalization rule of algorithms ($\mathsf{computeIter}$ vs. $\mathsf{ComputeIter}$, etc.)

    \end{itemize}
\end{itemize}
}
\section{Omitted Details of Section~\ref{sect:preliminaries}}\label{append:prelim}
\paragraph{Formal definition of $\mathsf{ComputeIter}$.} 
$\mathsf{ComputeIter}(p_L(\Vec{c}), p_U(\Vec{c}))$ returns a sufficient number $t$ of iterations of core counter {\sf ApproxMC6Core} such that the entire algorithm is $(\varepsilon, \delta)$-correct, provided the values of $p_L(\Vec{c})$ and $p_U(\Vec{c})$.
Specifically, $t$ is the minimum odd integer such that
\[
        \sum_{i=\lceil\frac{t}{2}\rceil}^{t} f(i,t,p_L(\Vec{c})) + \sum_{i=\lceil\frac{t}{2}\rceil}^{t} f(i,t,p_U(\Vec{c}))
        \leq \delta
        ,
\]
where $f(i,t,p):=\binom{t}{i}(p)^{i}(1-p)^{t-i}$ is the probability of getting exactly $i$ successes in $t$ independent Bernoulli trials (with the same rate $p$).

\section{Omitted Details of Section~\ref{sect:formalizeIntoOptimization}}\label{append:formalizeIntoOptimization}

\begin{mylemma}[\cite{YangM23}]\label{lem:concentration} 
    Let $F$ be a formula with $|\Vars(F)| = n$. 
    For every $0<\beta<1$, $\gamma >1$, and $1 \leq m \leq n$, we have the following:
    \begin{enumerate}
        \item $\MyPr[\Cnt{F}{m} \leq \beta \sfE[\Cnt{F}{m}]] \leq \frac{1}{1+ (1-\beta)^2 \sfE[\Cnt{F}{m}]} $,
        \item $\MyPr[\Cnt{F}{m} \geq \gamma \sfE[\Cnt{F}{m}]] \leq \frac{1}{1+ (\gamma - 1)^2 \sfE[\Cnt{F}{m}]} $. \qed
    \end{enumerate}
\end{mylemma}

\section{Proofs}\label{append:proofs}
\subsection{Proof of Lemma~\ref{lem:ConcreteConcentration}}\label{appendLem:ConcreteConcentration} 
\textbf{Lemma~\ref{lem:ConcreteConcentration}.} 
\textit{
    For given $\thresh >0$ and $\varepsilon >0$, let
    \begin{align*}
    q_T(a,k) &= 
        \begin{cases}
        \frac{1}{1+ (1-\frac{1}{(a \times 2^{k-1})})^2 \times a \times 2^{k-1} \times \thresh}        & \text{if } a \times 2^{k-1} > 1,\\  
        1    & \text{otherwise},
    \end{cases}\\
    q_L(a,k) &= \frac{1}{1+ (1-\frac{1}{(1+\varepsilon)})^2 \times a \times 2^{k-1} \times \thresh}, \\
    q_U(a,k) &= \frac{1}{1+ \varepsilon^2 \times a \times 2^{k-1} \times \thresh}, \\
    q_{\overline{T}\cup U}(a,k) &= 
        \begin{cases}
            \max\bigl\{
                \frac{1}{1+ (\frac{1}{(a \times 2^{k})}-1)^2 \times a \times 2^{k-1} \times \thresh}, q_U(a,k)
            \bigr\}        & \text{if } a \times 2^{k} < 1,\\  
            1    & \text{otherwise}.
        \end{cases}\\
    \end{align*}
    Then for any formula $F$,  $k\in \bbZ$, $Q \in \{L, U\}$, and $a_Q >0$ (which in turn specify $m_Q^*$), we have the following under $F,$ $\thresh$, and $\varepsilon$ (and independent of $\rnd$):
    \begin{align*}
        \MyPr[T_{m_Q^* - k}] &\leq q_T(a_Q, k),& \MyPr[\Lcnt{m_Q^* - k}] &\leq q_L(a_Q, k), \\
        \MyPr[\Ucnt{m_Q^* - k}] &\leq q_U(a_Q, k),& \MyPr[\overline{T_{m_Q^* - k}} \cup \Ucnt{m_Q^* - k}] &\leq q_{\overline{T}\cup U}(a_Q, k). 
    \end{align*}
}
\begin{proof}
For a fixed $\thresh >0$ and $\varepsilon>0$, take any formula $F$ with $|\Vars(F)| = n$, $k\in \bbZ$, $Q\in \{L,U\}$, and $a_Q > 0$. Observe that $\thresh$, $F$, $Q$, and $a_Q$ determine the value of $m_Q^*$; it is the smallest $m \in \bbZ$ that satisfies $2^{-m} \times |\sol(F)| < a_Q \times \thresh$. 
If $m_Q^*-k \not \in \{1, \ldots, n\}$, then all events considered in the lemma are empty, so the inequalities are trivially true. Thus we assume 
$m_Q^*-k \in \{1, \ldots, n\}$ below. 

We recall the following relationships that we frequently use in the proof:
For each $i \in \{1,\ldots, n\}$ and $Q \in \{L,U\}$,
\begin{align}
        \sfE[\cnt_i] = 2^{-i} \times |\sol|
    \quad, \quad %
    2^{-m_Q^*} \times |\sol| 
    < a_Q \times \thresh 
    \leq 2^{-(m_Q^*-1)} \times |\sol|. \label{eq:frequentlyUsed}
\end{align}

We prove $\MyPr[T_{m_Q^* - k}] \leq q_T(a_Q, k)$ as follows. We first observe
\begin{align}
    \thresh \leq \frac{1}{a_Q}\times 2^{-(m_Q^*-1)}
    = \frac{1}{a_Q \times 2^{k-1}}\times \sfE[\cnt_{m_Q^*-k}].\label{eq:ineq1}
\end{align}
By this we have
\begin{align*}
    \MyPr[T_{m_Q^*-k}] 
          &= \MyPr[\cnt_{m_Q^*-k} < \thresh] \\
          &\leq \MyPr
          \biggl[\cnt_{m_Q^*-k} < \frac{1}{a_Q \times 2^{k-1}}\times \sfE[\cnt_{m_Q^*-k}]
          \biggr].
\end{align*}
Hence we have the following, provided that $\frac{1}{a_Q \times 2^{k-1}} <1$ holds (for Lemma~\ref{lem:concentration}):
    \begin{align*}
        \MyPr[T_{m_Q^*-k}] 
          &\leq \frac{1}{1+ (1-\frac{1}{(a_Q \times 2^{k-1})})^2 \times \sfE[\cnt_{m_Q^*-k}]} &&\mbox{(Lemma~\ref{lem:concentration})} \\
          &\leq \frac{1}{1+ (1-\frac{1}{(a_Q \times 2^{k-1})})^2 \times a_Q \times 2^{k-1} \times \thresh} &&\mbox{(Ineq. (\ref{eq:ineq1}))}.
    \end{align*}
Similarly, we have $\MyPr[L_{m_Q^* - k}] \leq q_L(a_Q, k)$ as follows:
    \begin{align*}
        \MyPr[L_{m_Q^*-k}] 
          &\leq \MyPr\biggl[\cnt_{m_Q^*-k} \leq \frac{1}{1+\varepsilon} \sfE[\cnt_{m_Q^*-k}]\biggr] \\
          &\leq  \frac{1}{1+ (1-\frac{1}{(1+\varepsilon)})^2 \times \sfE[\cnt_{m_Q^*-k}]} &&\mbox{(Lemma~\ref{lem:concentration})}\\
          &\leq  \frac{1}{1+ (1-\frac{1}{(1+\varepsilon)})^2 \times a_Q \times 2^{k-1} \times \thresh} &&\mbox{(Ineq. (\ref{eq:ineq1}))}.
    \end{align*} 
Similarly, we have $\MyPr[U_{m_Q^* - k}] \leq q_U(a_Q, k)$ as follows:
    \begin{align*}
        \MyPr[U_{m_Q^*-k}] 
          &\leq \MyPr\bigl[\cnt_{m_Q^*-k} \geq (1+\varepsilon) \sfE[\cnt_{m_Q^*-k}]\bigr] \\
          &\leq  \frac{1}{1+((1+\varepsilon)-1)^2 \times \sfE[\cnt_{m_Q^*-k}]} &&\mbox{(Lemma~\ref{lem:concentration})}\\
          &\leq  \frac{1}{1+ \varepsilon^2 \times a_Q \times 2^{k-1} \times \thresh} &&\mbox{(Ineq. (\ref{eq:ineq1}))}.
    \end{align*} 
Finally, we prove $\MyPr[\overline{T_{m_Q^* - k}} \lor U_{m_Q^* - k}] \leq q_{\overline{T}\lor U}(a_Q, k)$ as follows. 
We observe 
\begin{align*}
    \thresh > \frac{1}{a_Q} \times 
    2^{-m_Q^*}
    = \frac{1}{a_Q \times 2^{k}}\times \sfE[\cnt_{m_Q^*-k}].
\end{align*}
By this we have
\begin{align*}
    \MyPr[\overline{T_{m_Q^* - k}}] 
          &= \MyPr[\cnt_{m_Q^*-k} \geq \thresh] \\
          &\leq \MyPr
          \biggl[\cnt_{m_Q^*-k} \geq \frac{1}{a_Q \times 2^{k}}\times \sfE[\cnt_{m_Q^*-k}]
          \biggr].
\end{align*}
Hence we have the following, provided that $\frac{1}{a_Q \times 2^{k}} >1$ holds (for Lemma~\ref{lem:concentration}):
    \begin{align*}
        \MyPr[\overline{T_{m_Q^* - k}}] 
          &\leq \frac{1}{1+ (\frac{1}{(a_Q \times 2^{k})}-1)^2 \times \sfE[\cnt_{m_Q^*-k}]} &&\mbox{(Lemma~\ref{lem:concentration})} \\
          &\leq \frac{1}{1+ (\frac{1}{(a_Q \times 2^{k})}-1)^2 \times a_Q \times 2^{k-1} \times \thresh} &&\mbox{(Ineq. (\ref{eq:ineq1}))}.
    \end{align*}
Notice that, given all the parameters fixed, we have either $\overline{T_{m_Q^* - k}} \subseteq U_{m_Q^* - k}$ or $U_{m_Q^* - k} \subseteq \overline{T_{m_Q^* - k}}$; hence $\MyPr[\overline{T_{m_Q^* - k}} \lor U_{m_Q^* - k}] = \max\{\MyPr[\overline{T_{m_Q^* - k}}], \MyPr[U_{m_Q^* - k}]  \}$, and we have what we need.
    \qed
    
\end{proof}

\subsection{Proof of Theorem~\ref{thm:soundness}}\label{appendThm:soundness}

\textbf{Theorem~\ref{thm:soundness} (soundness of $p_L(\Vec{c})$ and $p_U(\Vec{c})$ as probability bounds).}
\textit{
    For a given $\varepsilon > 0$, 
    suppose $\Vec{c} = (\thresh,\rnd,a_L, k_L^\downarrow, k_L^\uparrow, a_U, k_U^\downarrow, k_U^\uparrow)$ is an $\varepsilon$-valid vector. 
    Then for an arbitrary CNF formula $F$, 
    the relationships (\ref{eq:cutoffArgumentL}) and (\ref{eq:cutoffArgumentU}) hold under $(F, \thresh, \rnd)$ and $\varepsilon$, and thus,  
    we have 
    $\MyPr[L] \leq p_L(\Vec{c})$ and $\MyPr[U] \leq p_U(\Vec{c})$ under $(F, \thresh, \rnd)$ and $\varepsilon$, 
    where $p_L(\Vec{c})$ and $p_U(\Vec{c})$ are as defined in (\ref{eq:DefOfProbBounds}).
}
\begin{proof}
\todo{(done) JJ: now the entire proof is to be moved to the appendix. update accordingly}It suffices to show, if the underlying vector $\Vec{c}$ is valid, then the cut-off arguments (\ref{eq:cutoffArgumentL}) and (\ref{eq:cutoffArgumentU}) are true under any $F$.
Once we show this, the theorem is immediately proved by Lemma~\ref{lem:ConcreteConcentration}.
We first prove the properties stated in Item~\ref{item:phi2},~\ref{item:phi3} and~\ref{item:phi4}. 
After that, 
we prove that the cut-off arguments (\ref{eq:cutoffArgumentL}) and (\ref{eq:cutoffArgumentU}) are true under any $F$
and a valid vector $\vec{c}$, by invoking the properties stated in Item~\ref{item:phi2},~\ref{item:phi3} and~\ref{item:phi4}. 
In what follows, assume $\varepsilon > 0$ and an $\varepsilon$-valid vector $\Vec{c}$ are fixed, and events are considered under an arbitrary $F$ with $|\Vars(F)| = n$.

\medskip

Item~\ref{item:phi2} claims $\Lout{i} = \emptyset$  for each  $i \geq m_L^*-k_L^\uparrow$.  
If $m_L^*-k_L^\uparrow > n$ then this is true by definition.  
    Otherwise, observe $\sfE[\cnt_{m_L^*-k_L^\uparrow}] \geq \sfE[\cnt_i] $ holds for each $m_L^*-k_L^\uparrow \leq i \leq n$; 
    thus it suffices to show $\Lrnd{m_L^*-k_L^\uparrow} = \emptyset$, i.e., $\rnd \geq \frac{1}{1+\varepsilon}\sfE[\cnt_{m_L^*-k_L^\uparrow}]$. Now observe $\varphi_2(\Vec{c})$ implies what we need, as follows: 
    \begin{align*}
        \rnd &\geq \frac{a_L}{1+\varepsilon}\times 2^{k_L^\uparrow} \times \thresh &&\mbox{(Def. of $\varphi_2$)} \\
        &> \frac{1}{1+\varepsilon}\times 2^{k_L^\uparrow} \times 2^{-m_L^*} \times |\sol| &&\mbox{(Eq.~\ref{eq:frequentlyUsed})} \\
        &= \frac{1}{1+\varepsilon}\sfE[\cnt_{m_L^*-k_L^\uparrow}] .&&\mbox{(Eq.~\ref{eq:frequentlyUsed})} 
    \end{align*}

Item~\ref{item:phi3} claims 
$\Urnd{i} = \emptyset$ for each $i \leq m_U^*-k_U^\uparrow$. 
If $m_U^*-k_U^\uparrow <1$ then this is true by definition.  
Otherwise, 
observe $\Urnd{i} = \emptyset$ iff 
$\rnd \leq (1+\varepsilon)\sfE[\cnt_{i}]$ for each $i \in \{1, \ldots, n \}$; 
also $\sfE[\cnt_{m_U^*-k_U^\uparrow}] \leq \sfE[\cnt_i] $ holds for each $1 \leq i \leq m_U^*-k_U^\uparrow$. 
Thus it suffices to show 
$\rnd \leq (1+\varepsilon)\sfE[\cnt_{m_U^*-k_U^\uparrow}]$. 
Now observe $\varphi_3(\vec{c})$ implies what we need, as follows: 
\begin{align*}
    \rnd &\leq (1+\varepsilon)a_U \times  2^{k_U^\uparrow -1} \times \thresh &&\mbox{(Def. of $\varphi_3$)} \\
    &\leq (1+\varepsilon) \times 2^{k_U^\uparrow -1} \times 2^{-(m_U^*-1)} \times |\sol| &&\mbox{(Eq.~\ref{eq:frequentlyUsed})} \\
    &=(1+\varepsilon)\sfE[\cnt_{m_U^*-k_U^\uparrow}]. &&\mbox{(Eq.~\ref{eq:frequentlyUsed})}
\end{align*}

Item~\ref{item:phi4} claims $\Uout{i} = \emptyset$ for each $i \leq m_U^*-k_U^\downarrow$. 
For each $i <1$ then this is true by definition. 
For $1 \leq i \leq m_U^*-k_U^\downarrow$, we first show that $\varphi_4(\vec{c})$ implies 
$T_i \cap \Ucnt{i} = \emptyset$ for each $1 \leq i \leq m_U^*-k_U^\downarrow$, as follows: Assuming $T_i$, we have
\begin{align*}
    \cnt_i 
    &<    \thresh &&\mbox{(Def. of $T_i$)} \\
    &\leq \frac{1}{a_U}  \times  2^{-(m_U^*-1)} \times |\sol| &&\mbox{(Eq.~\ref{eq:frequentlyUsed})} \\
    &\leq (1+\varepsilon) \times 2^{(k_U^\downarrow -1)}  \times  2^{-(m_U^*-1)} \times |\sol| &&\mbox{(Def. of $\varphi_4$)} \\
    &\leq (1+\varepsilon) \times 2^{-i} \times |\sol| &&\mbox{($i \leq m_U^*-k_U^\downarrow$)} \\
    &=    (1+\varepsilon) \times \sfE[\cnt_i], &&\mbox{(Eq.~\ref{eq:frequentlyUsed})}   
\end{align*}
which means $\overline{\Ucnt{i}}$. Now, by $\varphi_1(\vec{c})$ and Item~\ref{item:phi3} we have $\Urnd{i} = \emptyset$ for each $1 \leq i \leq m_U^*-k_U^\downarrow$; thus we have $\Uout{i} = T_i \cap \overline{T_{i-1}} \cap \Ucnt{i}$ for such $i$, by definition of $\Uout{i}$; 
and hence $\Uout{i} = \emptyset$  for such $i$, by  $T_i \cap \Ucnt{i} = \emptyset$. 

\medskip

At this point, we have proven the properties described in Item~\ref{item:phi2},~\ref{item:phi3} and~\ref{item:phi4}.
Next, we prove that the cut-off arguments (\ref{eq:cutoffArgumentL}) and (\ref{eq:cutoffArgumentU}) are true under any $F$
and a valid vector $\vec{c}$.

For (\ref{eq:cutoffArgumentL}), 
we have $\Lout{i} \subseteq T_i$ and $T_i \subseteq T_{i+1}$ for each $i \in \bbZ$, by definition. 
Thus we have $\Lout{i} \subseteq T_{m_L^*-k_L^\downarrow}$ for each $i \leq m_L^*-k_L^\downarrow$.
Together with Item~\ref{item:phi2} above, we have $L \subseteq T_{m_L^*-k_L^\downarrow} 
\cup
\Lout{m_L^*-(k_L^\downarrow-1)} 
\cup \cdots \cup 
\Lout{m_L^*-(k_L^\uparrow+1)}$. 
As $\Lout{i} \subseteq \Lcnt{i}$ holds for any $i$, the cutoff argument (\ref{eq:cutoffArgumentL}) holds.

For (\ref{eq:cutoffArgumentU}), 
we have $\Uout{i} \subseteq \overline{T_{i-1}}$ and $\overline{T_i} \subseteq \overline{T_{i-1}}$ for each $i \in \bbZ$, by definition. 
Thus we have 
$\Uout{i}\subseteq \overline{T_{m_U^*-k_U^\uparrow}}$ 
For each $i > m_U^*-k_U^\uparrow$. 
Also Item~\ref{item:phi3} implies $\Uout{i}  \subseteq  \Ucnt{i}$ for each $i \leq m_U^*-k_U^\uparrow$. 
Thus together with Item~\ref{item:phi4} we have $U \subseteq \text{(RHS of (\ref{eq:cutoffArgumentU}))} \cup (\overline{T_n} \cap \hat{U})$. 
Now, if $m_U^* - k_U^\uparrow \leq n$, then  $\overline{T_n} \cap \hat{U} \subseteq \overline{T_n} \subseteq \overline{T_{m_U^*-k_U^\uparrow}}$; 
or otherwise we have 
$\Urnd{n} = \emptyset$ by Item~\ref{item:phi3}, and thus $2^n \leq \rnd \times 2^n \leq (1 +\varepsilon) |\sol|$, which says $\hat{U} = \emptyset$. 
Hence $\overline{T_n} \cap \hat{U} \subseteq \text{(RHS of (\ref{eq:cutoffArgumentU}))} $, and %
(\ref{eq:cutoffArgumentU}) holds. \qed
\end{proof}

\subsection{Proof of Theorem~\ref{thm:NarrowingDownTheSearchSpace}}\label{appendThm:NarrowingDownTheSearchSpace}
\textbf{Theorem~\ref{thm:NarrowingDownTheSearchSpace} (search space shrinking).}
\textit{
    Suppose $\sf{Obj}$ satisfies Assumption~\ref{ourAssumption}.
    For a given $\varepsilon>0$, consider the following constraints on $\Vec{c} \in \calC$: 
    \begin{align*}
    \psi_1(\Vec{c}) &\equiv k_L^\uparrow = 0 \land k_U^\downarrow = 1 \land  k_U^\uparrow = 0, &
    \psi_2(\Vec{c}) &\equiv k_L^\downarrow = 1  \lor k_L^\downarrow = 2, \\
    \psi_3(\Vec{c}) &\equiv a_L = \frac{(1+\varepsilon)^2}{2}a_U, &
    \psi_4(\Vec{c}) &\equiv \rnd = \frac{(1+\varepsilon)a_U}{2} \times \thresh.
\end{align*}
Then there exists a solution $\Vec{c}_{\text{sol}}$ to the optimization problem (\ref{line:ourOptimizationProb}) with the same $\varepsilon$ that satisfies $\psi_1(\Vec{c}_{\text{sol}})\land \psi_2(\Vec{c}_{\text{sol}})\land \psi_3(\Vec{c}_{\text{sol}})\land \psi_4(\Vec{c}_{\text{sol}})$. 
}

\todo{JJ: now the entire proof is to be moved to the appendix. update accordingly}

\begin{proof}
We say $\Vec{d} \in \calC$ is \emph{better than $\Vec{c} \in \calC$} 
if \begin{inparaenum}[(a)]
\item $\Vec{d}$ is valid if $\vec{c}$ is, 
\item $p_L(\Vec{c}) \geq p_L(\Vec{d})$ and $p_U(\Vec{c}) \geq p_U(\Vec{d})$, and
\item $\Vec{c}$ and $\Vec{d}$ have the same value of $\thresh$. 
\end{inparaenum}
We say $\Vec{c}$ and $\Vec{d}$ are \emph{equivalent} if they are better than each other. 
Observe, under Assumption~\ref{ourAssumption}, a better vector gives a smaller value of $\mathsf{Obj}$ when it is fed as an input. 
Thus it suffices to show that, for any valid vector $\Vec{c}$, 
there exists $\Vec{d}$ that is better than $\Vec{c}$ and satisfies  
$\psi_1(\vec{d})\land \psi_2(\vec{d})\land \psi_3(\vec{d})\land \psi_4(\vec{d})$.

Before showing the existence of $\vec{d}$, we observe a certain monotonic behavior of $p_L$, $p_U$, and their components (cf. Lemma~\ref{lem:ConcreteConcentration}). 
\begin{itemize}
    \item $q_T(a,k)$ and $q_L(a,k)$ are decreasing with respect to $a$. 
    Thus $p_L(\vec{c})$ is decreasing with respect to $a_L$ in $\vec{c}$. 
    \item Let us write $q_{\overline{T}\lor U}(a)$ to denote $q_{\overline{T}\lor U}(a,0)$. Then $q_{\overline{T}\lor U}(a)$ is a continuous function that is decreasing over $a \in (0,\frac{1}{1+\varepsilon}]$, and increasing over $a \in (\frac{1}{1+ \varepsilon},\infty)$. 
    Thus $a= \frac{1}{1+ \varepsilon}$ gives the smallest value to $q_{\overline{T}\lor U}(a)$. 
\end{itemize}
\medskip
    \noindent
    $\mathbf{(k_L^\uparrow = k_U^\uparrow = 0.)}$ 
    We first observe that we can use a fixed value for either $k_Q^\downarrow$ or $k_Q^\uparrow$ for each $Q \in \{L,U\}$.   
    This is based on an observation that, 
    if we modify a valid vector $\Vec{c}$ into $\Vec{d}$ by letting $a_Q := a_Q \times 2^k$, $k_Q^\downarrow := k_Q^\downarrow -k$ and $k_Q^\uparrow := k_Q^\uparrow -k$, then $\Vec{c}$ and $\Vec{d}$ are equivalent.
    Roughly speaking, this is because substituting $a_Q$ with $a_Q \times 2^k$ is equivalent to decrementing $k_Q^\downarrow$ and $k_Q^\uparrow$ by $k$. 
    Thus we can assume $k_L^\uparrow = k_U^\uparrow = 0$.
    
    The formal proof is as follows.
    Take any $x \in \bbZ$ and let $\Vec{c} = (\thresh, \rnd, a_L, k_L^\downarrow, k_L^\uparrow, a_U, $ $k_U^\downarrow, k_U^\uparrow)$ be given. 
    Let $x' = k_L^\uparrow - x$, and let $\Vec{d} = (\thresh, \rnd, a_L\times 2^{x'}, k_L^\downarrow-x', x, a_U, k_U^\downarrow, $ $k_U^\uparrow)$. 
    We show $\vec{c}$ and $\vec{d}$ are equivalent, 
    thus in particular, $\vec{c}$ and $\vec{d_0} = (\thresh, \rnd, a_L\times 2^{k_L^\downarrow}, k_L^\downarrow - k_L^\uparrow, 0, a_U, k_U^\downarrow, $ $k_U^\uparrow)$ are equivalent.
    The proof is done 
    as follows:  
    we have $\varphi_1(\Vec{c})$ iff $\varphi_1(\Vec{d})$
    by 
    $k_L^\downarrow > k_L^\uparrow$ iff $k_L^\downarrow -x' > k_L^\uparrow - x' = x$; 
    we have $\varphi_2(\Vec{c})$ iff $\varphi_2(\Vec{d})$
    by
    $a_L \times 2^{k_L^\uparrow } 
    = (a_L \times 2^{x'})\times 2^{k_L^\uparrow -x'} 
    = (a_L \times 2^{x'})\times 2^x $; 
    and clearly $\varphi_3(\Vec{c}) \land \varphi_4(\Vec{c})$ holds iff $\varphi_3(\Vec{d})\land \varphi_4(\Vec{d})$ does, because $\varphi_3$ and $\varphi_4$ only involve parameters that are unchanged in $\Vec{d}$. 
    It is easy to check from the definition that $p_L(\Vec{c}) = p_L(\Vec{d})$ holds; we obviously have $p_U(\Vec{c}) = p_U(\Vec{d})$ too, as the relevant parameters in $\Vec{c}$ and $\Vec{d}$ are the same. 
    In the same vain, we can show that $\vec{c}$ and 
    $(\thresh, \rnd, a_L\times 2^{k_L^\downarrow}, k_L^\downarrow - k_L^\uparrow, 0, a_U\times 2^{k_U^\downarrow}, k_U^\downarrow - k_U^\uparrow, 0)$ are equivalent. 
    Thus, we can assume $k_L^\uparrow = k_U^\uparrow = 0$.

    \medskip 
    \noindent
    $\mathbf{(\psi_3 \land \psi_4, k_U^\downarrow = 1.)}$ 
    Next, we show that we can assume $\psi_3(\Vec{c}) \land \psi_4(\Vec{c})$ and $k_U^\downarrow = 1$.
    We first give the proof sketch and then provide a formal proof.
    Observe that $\varphi_2(\Vec{c}) \land \varphi_3(\Vec{c})$ now looks like the following: 
    \begin{align}
        \frac{a_L}{1+\varepsilon}\times \thresh \leq 
     \rnd \leq \frac{(1+\varepsilon)a_U}{2} \times \thresh.\label{eq:phi2andPhi3_}
    \end{align}
    Now, for a given $\Vec{c}$, consider increasing the value of $a_L$ and $\rnd$ so that the inequalities in~(\ref{eq:phi2andPhi3_}) become equations. 
    It turns out that such a modification makes $\Vec{c}$ better; thus we can assume $\psi_3(\Vec{c}) \land \psi_4(\Vec{c})$.
    Next, consider modifying $\Vec{c}$ by letting $k_U^\downarrow := 1$ and $a_U := \max\{a_U, \frac{1}{1+\varepsilon}\}$, while also updating $a_L$ and $\rnd$ accordingly. 
    Here, the update of $a_U$ is necessary to satisfy $\varphi_4$, which now requires $a_U \geq \frac{1}{1+\varepsilon}$.
    Again, such a modification makes $\Vec{c}$ better; as we already assumed $k_L^\uparrow = k_U^\uparrow = 0$, 
    we can now assume $\psi_1(\Vec{c})$. 

    The formal proof is as follows.
    For a given  $\Vec{c} = (\thresh, \rnd, a_L, k_L^\downarrow, 0, a_U, $ $k_U^\downarrow, 0)$, let $\vec{d} = (\thresh, \rnd', $ $ a'_L, k_L^\downarrow, 0, a'_U, 1, 0)$, 
    where 
    $a'_U = \max\{ \frac{1}{1+\varepsilon}, a_U\}$, $\rnd' = \frac{1+\varepsilon}{2} \times a_U' \times \thresh$, and 
    $a'_L = \frac{(1+\varepsilon)^2}{2} a_U'$. 
    We show $\vec{d}$ is better than $\vec{c}$. 

     The validity of $\Vec{d}$ is easy to check. 
    We have $a_L' \geq a_L$ by $a_L' = \frac{(1+\varepsilon)^2}{2} a_U' \geq \frac{(1+\varepsilon)^2}{2} a_U \geq a_L$ (the last inequality is derived from $\varphi_2(\Vec{c}) \land \varphi_3(\Vec{c})$), and thus we have $p_L(\Vec{d}) \leq p_L(\Vec{c})$ by monotonicity of $q_T$ and $q_L$ (and by the fact that $\Vec{c}$ and $\Vec{d}$ have the same cut-off points for $L$). 
    To prove $p_U(\Vec{c}) \geq p_U(\Vec{d})$, observe %
    \begin{align*}
        p_U(\Vec{c}) &= q_U(a_U, k_U^\downarrow-1) + \ldots + q_U(a_U, 1) + q_{\overline{T}\lor U}(a_U) &&\mbox{(Def. of $p_U$)} \\
        &\geq q_{\overline{T}\lor U}(a_U)  && \\
        &\geq q_{\overline{T}\lor U}(\frac{1}{1+\varepsilon})  && \mbox{($\frac{1}{1+\varepsilon}$ minimizes $q_{\overline{T}\lor U}$)}.
    \end{align*}
    As we have either $p_U(\Vec{d}) = q_{\overline{T}\lor U}(a_U)$ or $p_U(\Vec{d}) = q_{\overline{T}\lor U}(\frac{1}{1+\varepsilon})$, the claim holds. 
    Thus, we can assume that $\psi_3(\Vec{c}) \land \psi_4(\Vec{c})$ and $k_U^\downarrow = 1$.
    
    \medskip
    \noindent
    $\mathbf{(\psi_2.)}$
    Finally, we can show that a modification $k_L^\downarrow := k_L^\downarrow -1$ makes $\Vec{c}$ better if $k_L^\downarrow \geq 3$; thus we can assume $k_L^\downarrow \leq 2$.  
    As we already assumed $k_L^\uparrow = 0$, we have $k_L^\downarrow \geq 1$ as a requirement from $\varphi_1$. Hence, we can now assume $\psi_2(\Vec{c})$.

    The formal proof is as follows.
    Let a valid vector $\Vec{c} = (\thresh, \rnd, a_L, k_L^\downarrow, 0, a_U, 1, 0)$ be given; by what we have shown so far, assume $\rnd = \frac{1+\varepsilon}{2} \times a_U \times \thresh$ and $a_L = \frac{(1+\varepsilon)^2}{2} a_U$  w.l.o.g. 
    We show, if $k_L^\downarrow \geq 3$, then $\Vec{d} = (\thresh, \rnd, a_L, k_L^\downarrow -1, 0, a_U, 1, 0)$ is better than $\vec{c}$. 

    The validity of $\Vec{d}$ and $p_U(\Vec{c}) = p_U(\Vec{d})$ are easy to check. For $p_L(\Vec{c}) \geq p_L(\Vec{d})$, observe
\begin{align}
    p_L (\Vec{c}) -p_L(\Vec{d}) 
    = q_T(a_L, k_L^\downarrow) + q_L(a_L, k_L^\downarrow-1) - q_T(a_L, k_L^\downarrow-1). \label{eq:checkIfDecrementingMakesItBetter}
\end{align}

Thus it suffices to show $q_L(a_L, k_L^\downarrow-1) - q_T(a_L, k_L^\downarrow-1) \geq 0$. Observe 
$q_L(a, k) - q_T(a, k)$ is of the form
\begin{align}
\frac{1}{1+ (1-\frac{1}{(1+\varepsilon)})^2 \times F(a,k)}
-
\frac{1}{1+ (1-\frac{1}{(a \times 2^{k-1})})^2 \times F(a,k)},
\end{align}
where $F(a,k) = a \times 2^{k-1} \times \thresh$. Hence we have $q_L(a, k) - q_T(a, k) \geq 0$ iff $2^k \geq \frac{2(1+\varepsilon)}{a}$. 
Now $\varphi_4(\Vec{c})$ requires $a_U \geq \frac{1}{1+\varepsilon}$, so by the validity of $\Vec{c}$ and the assumption $a_L = \frac{(1+\varepsilon)^2}{2} a_U$, we have $a_L \geq \frac{1+\varepsilon}{2}$; 
thus we have $\frac{2(1+\varepsilon)}{a_L} \leq 4 \leq 2^{k_L^\downarrow -1}$ (recall $k_L^\downarrow \geq 3$). 
Hence we have what we need. \qed
    
\end{proof}

\subsection{Proof of Corollary~\ref{cor:reducedOptProb}}\label{append:reducedOptProb}
\textbf{Corollary~\ref{cor:reducedOptProb} (a reduced problem of~(\ref{line:ourOptimizationProb})}
\textit{
    For given $\thresh$, $a_U$ and $k_L^\downarrow$, define \todo{[maybe]Probably we can use concrete description or $\hat{p}_L$ etc. at this point, it may make the presentation simpler}
    $\Vec{c}(\thresh,a_U,k_L^\downarrow)$ by
    \[
    \Vec{c}(\thresh,a_U,k_L^\downarrow) = 
    (\thresh, \frac{(1+\varepsilon)a_U}{2} \times \thresh,\frac{(1+\varepsilon)^2}{2}a_U,k_L^\downarrow,0,a_U,1,0),
    \]
    and let $\hat{p}_Q(\thresh, a_U) = \min\{p_Q(\Vec{c}(\thresh,a_U,1)),p_Q(\Vec{c}(\thresh,a_U,2))\}$, where $Q\in\{L,U\}$. Then for $\mathsf{Obj}$ that satisfies  Assumption~\ref{ourAssumption}, the global minimum of the following optimization problem coincides with that of (\ref{line:ourOptimizationProb}):
\begin{align*}
\begin{split}
    \underset{(\thresh, a_U)}{\mbox{\bf Minimize}}
    &\quad \mathsf{Obj}(\hat{p}_L(\thresh, a_U), \hat{p}_U(\thresh, a_U), \thresh) \\
    \mbox{\bf Subject to}
    &\quad
    \thresh \geq 2, 
     \quad 
    a_U \geq \frac{1}{1+\varepsilon}. 
\end{split}
\tag{\ref{ourReducedOptimizationProb}}
\end{align*}
Furthermore, if $(\thresh, a_U)$ is a solution to~(\ref{ourReducedOptimizationProb}), and  
$k \in \{1,2\}$ is the number 
such that $\hat{p}_L(\thresh, a_U) = p_L(\Vec{c}(\thresh,a_U,k))$, 
then $\Vec{c}(\thresh,a_U,k)$ is a solution to~(\ref{line:ourOptimizationProb}). 
}
\begin{proof}
    In this proof, slightly abusing the notation, we occasionally write $\mathsf{Obj}(\Vec{c})$ to denote $\mathsf{Obj}(p_L(\Vec{c}), p_U(\Vec{c}), \thresh)$. 
    
    Let $X$ be the set of all tuples $(\thresh, a_U, k_L^\downarrow)$ such that $a_U \geq \frac{1}{1+\varepsilon}$ and $k_L^\downarrow \in \{1,2\}$; 
    also let $Y$ be the set of all valid vectors $\Vec{c}$ that also satisfy $\psi_1 (\Vec{c})\land \psi_2 (\Vec{c})\land \psi_3 (\Vec{c})\land \psi_4(\Vec{c})$. 
    It is easy to see $(\thresh, a_U, k_L^\downarrow) \mapsto \Vec{c}(\thresh, a_U, k_L^\downarrow)$ is a bijection from $X$ to $Y$ and, for each $(\thresh, a_U)$ with  $a_U \geq \frac{1}{1+\varepsilon}$,%
\begin{align}
        \mathsf{Obj}(\hat{p}_L(\thresh, a_U), \hat{p}_U(\thresh, a_U), \thresh) = \min_{k \in \{1,2\}}\mathsf{Obj}(\Vec{c}(\thresh, a_U, k)).\label{eq:inCorollary1}
\end{align}
    Here, notice we have $p_U(\Vec{c}(\thresh,a_U,1)) = p_U(\Vec{c}(\thresh,a_U,2))$; so the $\min$ operation and $\mathsf{Obj}$ commute in the LHS. Hence, by letting $\widehat{\mathsf{Obj}}(\thresh,a_U) = \mathsf{Obj}(\hat{p}_L(\thresh, a_U), \hat{p}_U(\thresh, a_U), \thresh)$, we have
    \begin{align*}
        \min_{(\thresh,a_U); a_U \geq\frac{1}{1+\varepsilon}} \widehat{\mathsf{Obj}}(\thresh,a_U) 
        &= \min_{(\thresh,a_U,k) \in X} \mathsf{Obj}(\Vec{c}(\thresh, a_U, k)) \\
        &= \min_{\Vec{c} \in Y}\mathsf{Obj}(\Vec{c}) \\
        &= \min_{\Vec{c}}\mathsf{Obj}(\Vec{c}).
    \end{align*}
    Here, the first equation is due to~(\ref{eq:inCorollary1}); 
    the second equation is due to the bijectivity of $(\thresh, a_U, k_L^\downarrow) \mapsto \Vec{c}(\thresh, a_U, k_L^\downarrow)$;
    and the third equation is due to Theorem~\ref{thm:NarrowingDownTheSearchSpace}. This proves the coincidence of the global minimum.
    
    Now let $(\thresh, a_U)$ be a solution to~(\ref{ourReducedOptimizationProb}), and  $k \in \{1,2\}$ be the number such that $\hat{p}_L(\thresh, a_U) = p_L(\Vec{c}(\thresh,a_U,k))$. 
    The equation~(\ref{eq:inCorollary1}) tells us 
    \[
    (\thresh,a_U,k) = \argmin_{(\thresh',a_U',k') \in X} \mathsf{Obj}(\Vec{c}(\thresh', a_U', k')),
    \]
    and by the above equations we know $\Vec{c}(\thresh,a_U,k) = \argmin_{\Vec{c}} \mathsf{Obj}(\Vec{c})$. \qed
    
\end{proof}

\newpage
\section{Supplemental Materials on Experiments}\label{append:experiments}

\paragraph{Ternary search algorithm for searching optimal $t$ (and the underlying $a_U$).}
\begin{algorithm}[t]
	\caption{{\sf TernarySearch}$(\thresh, \varepsilon, \delta)$}
      \label{alg:TernarySearch}
	\begin{algorithmic}[1]
        \State $L\gets \frac{1}{1+\varepsilon}$; $R\gets 1$;
        \State $\mathsf{absPrecision}\gets 10^{-3}$;
        \While{$\mathrm{abs}(R-L)\ge \mathsf{absPrecision}$}
            \State $L'\gets L+(R-L)/3$;
            \State $R'\gets R - (R-L)/3$;
            \If {$\mathsf{Calc\_t}(\thresh, L', \delta) > \mathsf{Calc\_t}(\thresh, R', \delta)$}
                \State $L\gets L'$;
            \Else
                \State $R\gets R'$;
            \EndIf
        \EndWhile
        \State $a_U\gets (L+R)/2$; $t\gets \mathsf{Calc\_t}(\thresh, a_U)$;
        \State \Return $(t, a_U)$;
	\end{algorithmic}
\end{algorithm}

For each fixed $\thresh$, we can perform a ternary search on $a_U$ (i.e., Algorithm~\ref{alg:TernarySearch}) to find the minimum $t$.
This is based on our experimental observation that $t$ exhibits unimodal behavior\footnote{
We performed an empoirical test %
with multiple $\varepsilon$, $\delta$, and $\thresh$ (e.g., $\varepsilon=0.8, \delta=0.001, 35\leq \thresh\leq 500$ and $\varepsilon=0.4, \delta=0.001, 35\leq \thresh\leq 500$). Within this, we didn't observe any non-unimodal behavior.} with respect to $a_U$ under a fixed $\thresh$ (see Figure~\ref{fig:unimodality}).
During the ternary search process, for fixed $\thresh$ and $a_U$, we employ subroutine {\sf Calc\_t} to determine the minimum $t$. Notably, within this subroutine, a binary search can be applied to find the minimum $t$, enabled by the monotonicity of the error rate with respect to $t$.

\paragraph{Definitions of standard metrics.} We give some omitted definitions here.

\begin{itemize}
    \item \textit{PAR-2 score.} 
    PAR-2 score is a way to measure how well an algorithm performs when solving a bunch of problems. 
    Our PAR-2 score is calculated as the mean of the runtime for solved cases and twice the time limit for unsolved cases. Recall that the time limit in our experiments is 5000 seconds.
    \item \textit{SpeedUp value.} For each instance $I$, the SpeedUp value is defined as the ratio of the runtime of $\mathsf{ApproxMC6}$ to the runtime of our $\mathsf{FlexMC}$. Specifically, it is given by $\frac{\tau_{base}(I)}{\tau_{our}(I)}$, where $\tau_{base}(I)$ and $\tau_{our}(I)$ are the runtimes of $\mathsf{ApproxMC6}$ and $\mathsf{FlexMC}$ on instance $I$, respectively.
    The overall SpeedUp is then computed as the geometric average of the SpeedUp values for all instances, excluding those instances where $\mathsf{FlexMC}$ solves within 1 second.

    \item \textit{Empirical error.} The empirical error for an instance $I$ is calculated as the absolute difference between the exact solution and the output of our algorithm, normalized by the exact solution. Specifically, it is given by $\frac{\mathsf{abs}(\mathsf{Exact}(I) - \mathsf{Output}(I))}{\mathsf{Exact}(I)}$, where $\mathsf{Exact}(I)$ and $\mathsf{Output}(I)$ are the exact solution of $I$ and the output of our algorithm on $I$, respectively. 
    The overall empirical error is then the mean of these individual empirical errors over all instances.
\end{itemize}

\begin{figure*}[!tb]
\centering
\begin{subfigure}[b]{0.48\textwidth}
\includegraphics[width=\textwidth]{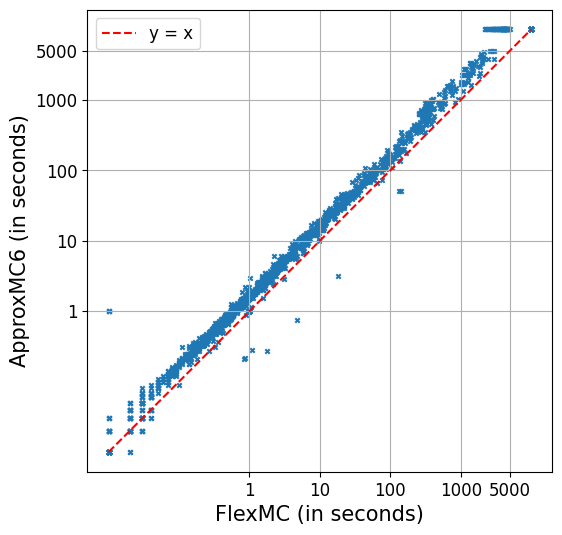}
\caption{$\varepsilon=0.8$, $\delta=0.001$.}
\end{subfigure}~~~~~~%
\begin{subfigure}[b]{0.48\textwidth}
\includegraphics[width=\textwidth]{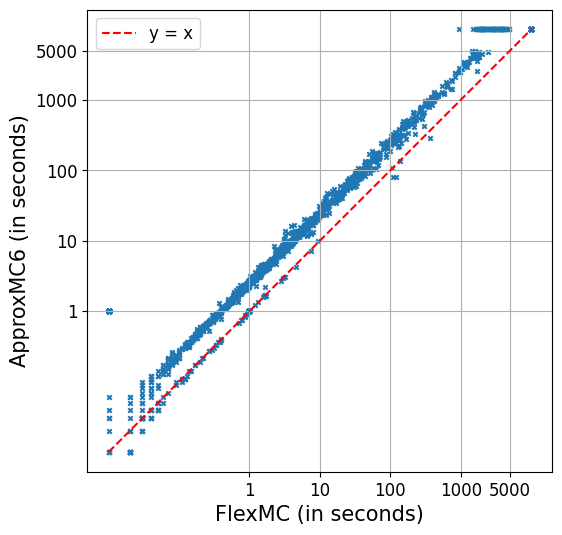}
\caption{$\varepsilon=0.4$, $\delta=0.001$.}
\end{subfigure}~~~~~~%
\caption{Comparison of runtimes of $\mathsf{ApproxMC6}$ and $\mathsf{FlexMC}$ in all instances.}\label{fig:scatterPlot}
\end{figure*}

\begin{figure}[htbp]
        \centering
        \includegraphics[width=1.05\linewidth]{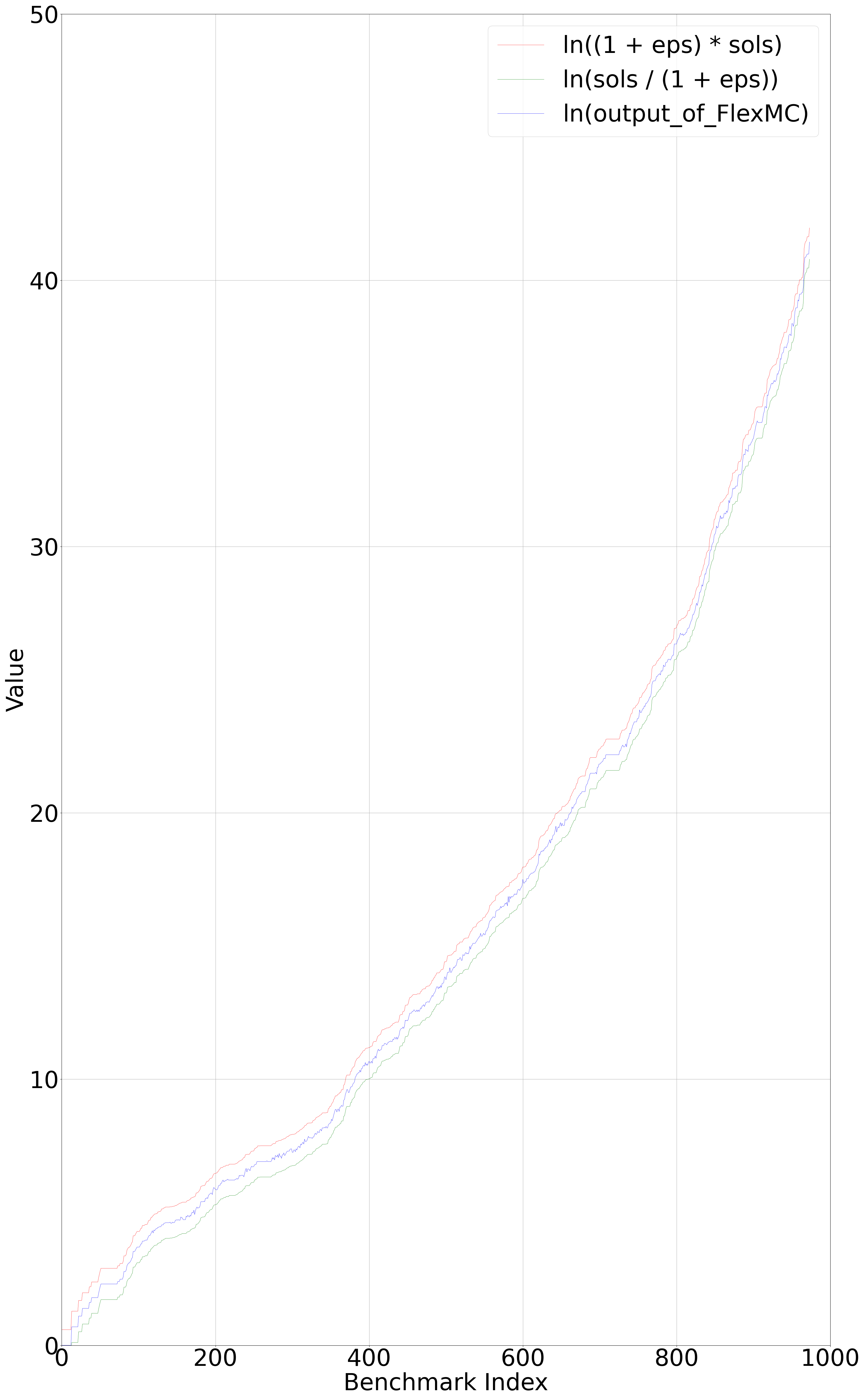}
        \caption{Benchmark-wise results in accuracy evaluation:  $\varepsilon$=0.8, $\delta$=0.001}
        \label{fig:accuracyBenchwise1}
    \end{figure}

\begin{figure}[htbp]
        \centering
        \includegraphics[width=1.05\linewidth]{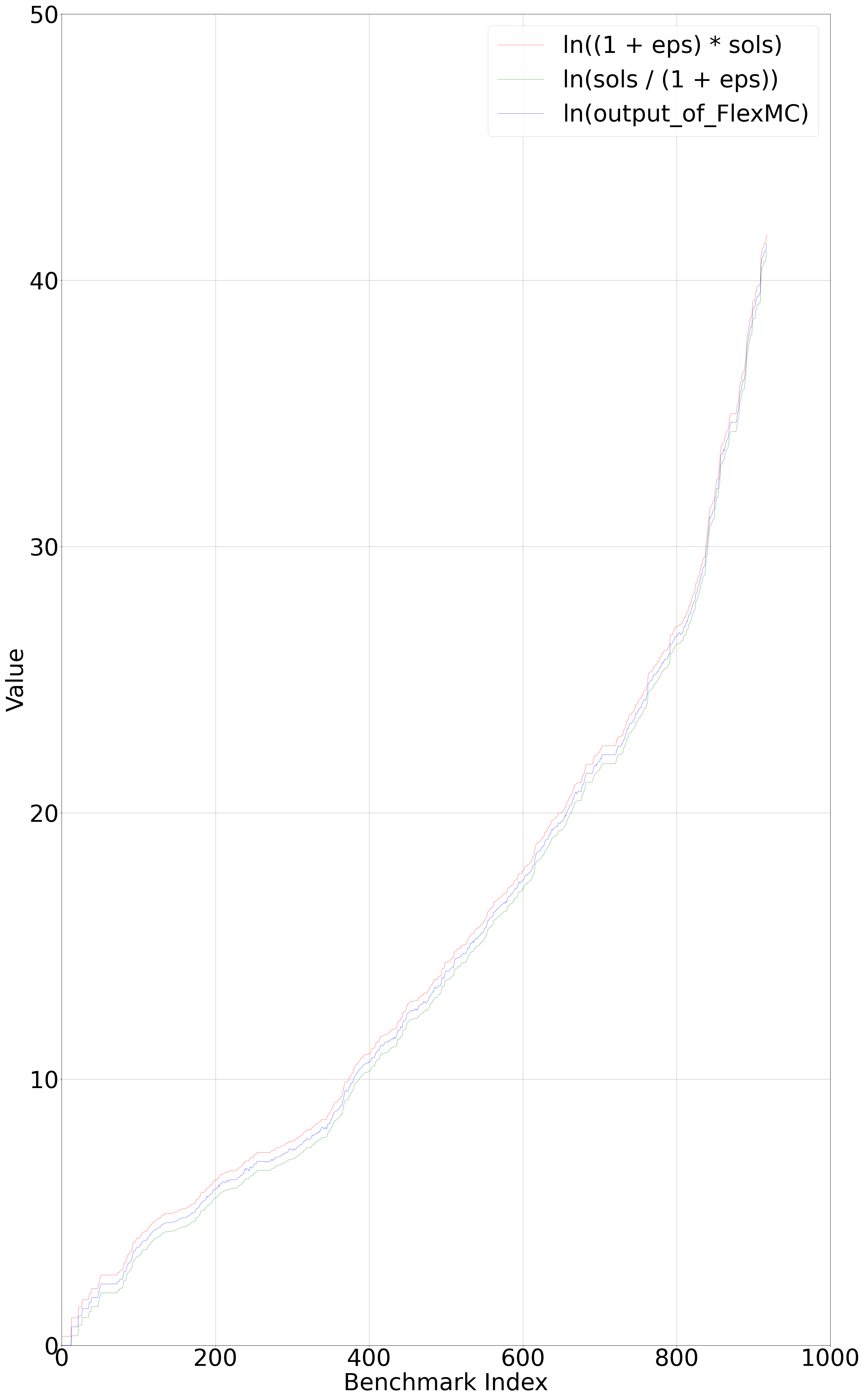}
        \caption{Benchmark-wise results in accuracy evaluation: $\varepsilon$=0.4, $\delta$=0.001}
        \label{fig:accuracyBenchwise2}
    \end{figure}

\paragraph{Additional experiment data.} 
The scatter plot of the runtime comparison between $\mathsf{FlexMC}$ and $\mathsf{ApproxMC6}$ is given in Figure~\ref{fig:scatterPlot}; 
the graph of benchmark-wise results in the accuracy evaluation of $\mathsf{FlexMC}$ is given in Figure~\ref{fig:accuracyBenchwise1} and Figure~\ref{fig:accuracyBenchwise2}.


\end{document}

%% file: Sec_Experiments.tex
\section{Experiments}
We performed experiments to examine the efficiency and accuracy of our updated approximate model counter, $\mathsf{FlexMC}$. 
Following the experiment design in the previous updates~\cite{YangM23,SoosGM20}, we consider two comparison targets: 
the latest version of $\mathsf{ApproxMC}$ to evaluate efficiency, 
and an exact model counter to evaluate accuracy. 
For the former, the current latest version is $\mathsf{ApproxMC6}$~\cite{YangM23}. 
For the latter, we use $\mathsf{Ganak}$~\cite{SRSM19}.
We note that an entry based on $\mathsf{ApproxMC6}$ won the Model Counting Competition 2024, justifying the use of $\mathsf{ApproxMC6}$ as the comparison target. 
We used a pre-processing tool $\mathsf{Arjun}$~\cite{soos2022arjun}. 

Recall both $\mathsf{FlexMC}$ and $\mathsf{ApproxMC6}$ are given as
Algorithm~\ref{alg:ApproxMC6}, 
where the only difference\footnote{
We ran experiment with $\varepsilon < 3$, and thus the difference in Footnote~\ref{footnote:alwaysRound} did not occur.
} 
between them is in Line~\ref{ln:initBegins}: 
While $\mathsf{FlexMC}$ computes $\mathsf{setParameters}(\varepsilon,\delta)$ by Algorithm~\ref{alg:FindOptPara}, 
$\mathsf{ApproxMC6}$ returns a precomputed value depending on the value of $\varepsilon$ (see~\cite{YangM23}).
We used a simple ternary search algorithm to realize the {\sf FindOptIter} procedure in Algorithm~\ref{alg:FindOptPara}, whose pseudocode is given in Appendix~\ref{append:experiments}. Runtime overhead by Algorithm~\ref{alg:FindOptPara} was less than 1 second. \todo{check}

We ran experiments with two choices of input parameters, namely $\varepsilon=0.8,\delta=0.001$ and $\varepsilon=0.4,\delta=0.001$. 
The first choice is the same as the one in the experiments for $\mathsf{ApproxMC6}$~\cite{YangM23}. 
We added the second choice for two reasons. 
First, for $\varepsilon = 0.4$, the gap of the bounds on $\MyPr[L]$ and $\MyPr[U]$ in $\mathsf{ApproxMC6}$ is much larger~\cite[Lemma 4]{YangM23}; 
based on the latest observation %
that the maximum of these two bounds is crucial for the performance~\cite[Lemma 3]{YangM23}, 
we postulate $\mathsf{FlexMC}$ has an additional advantage upon $\mathsf{ApproxMC6}$ due to its ability of balancing $p_L(\Vec{c})$ and $p_U(\Vec{c})$ (cf.~\S\ref{subsect:landscape}).
Second, we would like to know how well our objective function evaluates the actual performance of the resulting model counters, and this arrangement gives us additional data.

Our benchmark includes the previous 1890 datasets \cite{jiong_yang_2023_7931193} plus additional 800 data sets from track 1 and track 3 of Model Counting Competitions 2023-2024 \cite{fichte_2023_10012864, fichte_2024_14249068}.
The experiment ran on the computer with Intel(R) Xeon(R) Gold 6226R CPU @ 2.90GHz featuring 2 $\times$ 16 real cores and 512 GB of RAM. Each instance ran on a single core with a time limit 5000 seconds. 

\medskip
Our Research questions are the following: 

\noindent
{\bf RQ1.} Does $\mathsf{FlexMC}$ improve the runtime performance of $\mathsf{ApproxMC6}$?

\noindent
{\bf RQ2.} Does $\mathsf{FlexMC}$ have additional advantage on the runtime performance over $\mathsf{ApproxMC6}$ when the latter uses probability bounds with a large gap?

\noindent
{\bf RQ3.} Does our objective function reflect the performance of model counters?

\noindent
{\bf RQ4.} How is the empirical accuracy of the model counts by $\mathsf{FlexMC}$?

\begin{table}[t]
\centering
\begin{tabular}{|c|r|c|c|c|c|c|c|c|c|c|}
\hline
$\varepsilon$         & \multicolumn{1}{c|}{Algorithm} & $p_L$  & $p_U$  & $t$  & $\mathsf{thresh}$ &  $\mathsf{Obj}$ & $\mathsf{Obj}$-ratio & \#Solved & PAR2 & Speedup \\ \hline
\hline
                      & $\mathsf{ApproxMC6}$ & 0.157 & 0.169 & 19 & 72     & 1368 & 2.02       & 1907     & 3006 & -       \\ \cline{2-11} 
\multirow{-2}{*}{0.8} & $\mathsf{FlexMC}$    & 0.135 & 0.119 & 13 & 52     & 676  & -          & 1964     & 2829 & 1.65    \\ \hline
                      & $\mathsf{ApproxMC6}$ & 0.262 & 0.169 & 37 & 155    & 5735 & 2.71       & 1826     & 3323 & -       \\ \cline{2-11} 
\multirow{-2}{*}{0.4} & $\mathsf{FlexMC}$    & 0.137 & 0.117 & 13 & 163    & 2119 & -          & 1893     & 3077 & 2.46    \\ \hline
\end{tabular}
\caption{Efficiency comparison between $\mathsf{FlexMC}$ and $\mathsf{ApproxMC6}$. The entries $p_L$, $p_U$, $t$ and $\mathsf{thresh}$ show the values in Line~\ref{ln:initBegins} of
Algorithm~\ref{alg:ApproxMC6}; $\mathsf{Obj}$ means $t \times \thresh$. %
}\label{table:efficiencyComparison}
\end{table}

\begin{figure*}[!tb]
\centering
\begin{subfigure}[b]{0.48\textwidth}
\includegraphics[width=\textwidth]{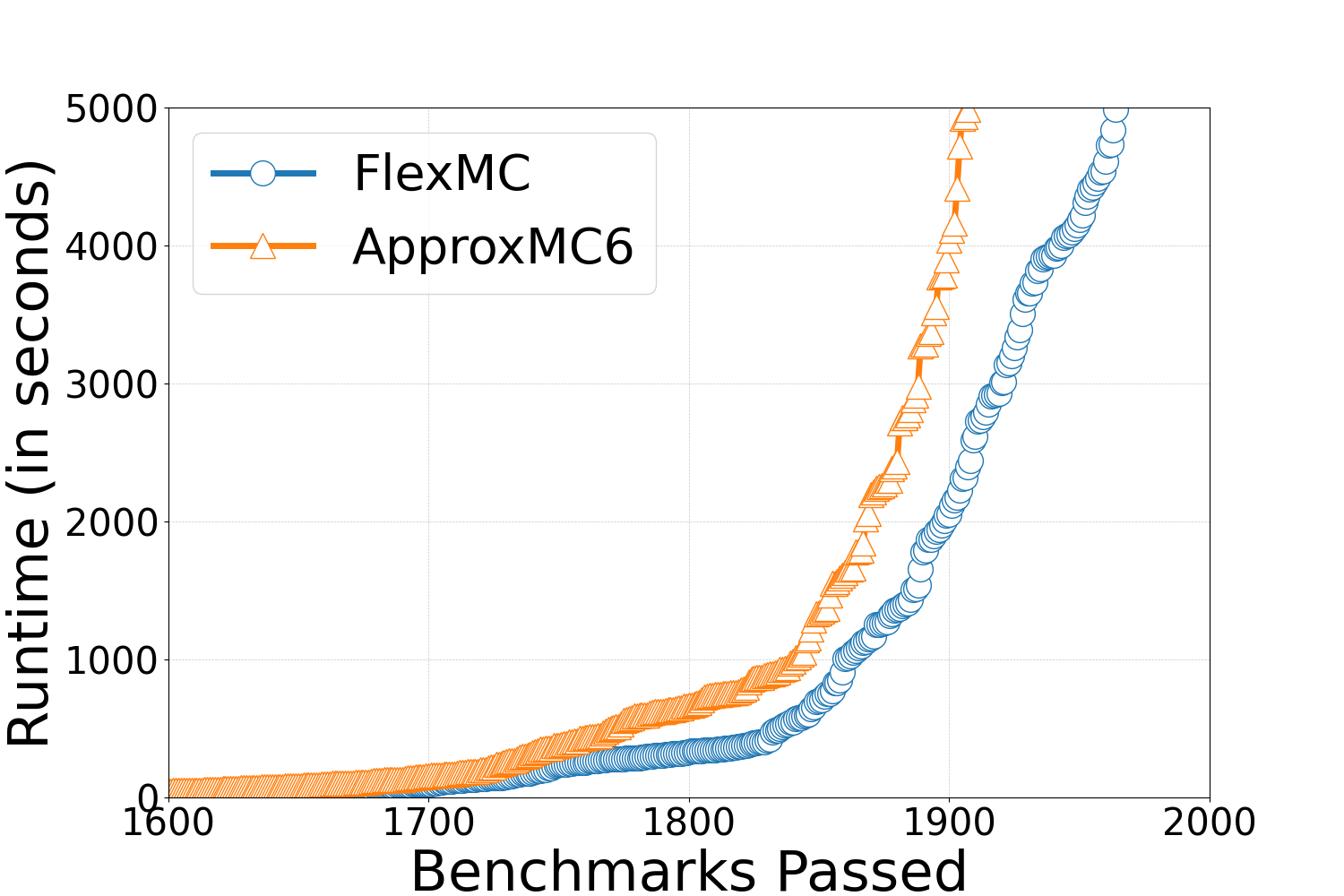}
\caption{$\varepsilon = 0.8, \delta = 0.001$}
\end{subfigure}~~~~~~%
\begin{subfigure}[b]{0.48\textwidth}
\includegraphics[width=\textwidth]{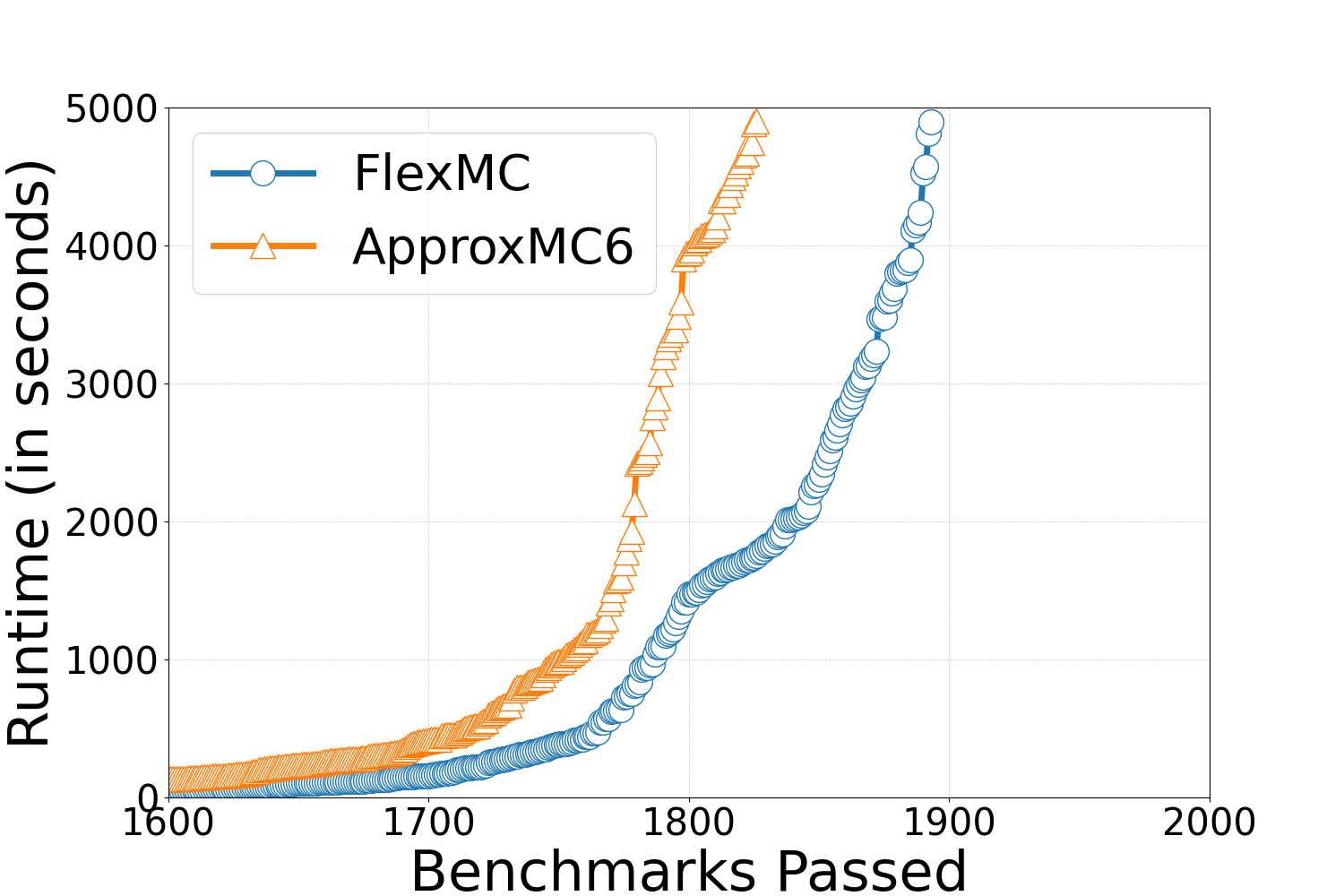}
\caption{$\varepsilon = 0.4, \delta = 0.001$}
\end{subfigure}~~~~~~%
\caption{Cactus plots showing behavior of $\mathsf{ApproxMC6}$ and $\mathsf{FlexMC}$}
\label{fig:cactusPlots}
\end{figure*}

\paragraph{Results 1: Efficiency evaluation.} 
Table~\ref{table:efficiencyComparison} 
summarizes the characteristics and performance of $\mathsf{ApproxMC6}$ and $\mathsf{FlexMC}$, including the number of solved instances, PAR-2 score, and speed-up rate. 
To reduce the noise of the constant runtime factor, the speed-up rate is computed excluding cases with runtime less than 1 second. 
The cactus plots are shown in Figure~\ref{fig:cactusPlots}. 
For definitions of some standard metrics in the results, e.g., the PAR-2 score, see Appendix~\ref{append:experiments}.\todo{We fill this}

On {\bf RQ1}, we observe nontrivial improvement of the efficiency in both cases of $\varepsilon$. 
Apart from a few singular cases, the speed-up ratio on individual benchmarks does not deviate too much from the mean value; the scatter plots are given in Appendix~\ref{append:experiments}.
This is a natural consequence because, at the code level, the main update in $\mathsf{FlexMC}$ appears in the use of different $t$ and $\mathsf{thresh}$. %

On {\bf RQ2}, 
$\mathsf{FlexMC}$ achieves a higher performance improvement under a smaller $\varepsilon$. 
We also observe $\mathsf{FlexMC}$ adopts more balanced values of $p_L$ and $p_U$ than those of $\mathsf{ApproxMC6}$. 
Meanwhile, we find yet another balancing phenomenon in the parameter choice in $\mathsf{FlexMC}$---that between the probability bounds and $\thresh$.
We observe the values of $p_L$ and $p_U$ themselves are much smaller than those of $\mathsf{ApproxMC6}$. 
This is realized by taking a larger value of $\mathsf{thresh}$; 
indeed, $\mathsf{thresh}$ in $\mathsf{FlexMC}$ is even larger than that in $\mathsf{ApproxMC6}$. 
This suggests that the flexibility of both $a_U$ and $\thresh$---as demonstrated in Figure~\ref{fig:errorBoundLandscape}---contribute to the performance improvement of $\mathsf{FlexMC}$. %

On {\bf RQ3}, We observe the speed-up ratio takes roughly from $80\%$ to $90\%$ of the value of $\mathsf{Obj}$-ratio. 
This is a natural consequence because the value of $\mathsf{Obj}_1^{(\delta)}$ is roughly proportional to the number of SAT calls by the model counters. This result also supports the adequacy of our objective function.

\begin{wrapfigure}[6]{r}{0.14\textwidth}
\vspace{-2em}
	\centering
\begin{tabular}{|c|c|}
\hline
$\varepsilon$ & $\varepsilon_{\mathrm{emp}}$ \\ \hline
0.8           & 0.027   \\ \hline
0.4           & 0.014   \\ \hline
\end{tabular}
\end{wrapfigure}

\paragraph{Results 2: Accuracy evaluation.} 
We provide the mean value of the empirical error rate $\varepsilon_{\mathrm{emp}}$ of the approximate count by $\mathsf{FlexMC}$ in the right table. 
The graph of benchmark-wise results is given in Appendix~\ref{append:experiments}. 
We observe the achieved error rate by $\mathsf{FlexMC}$ is much smaller than the theoretical error tolerance.

%% file: Sec_Related_Works.tex
\section{Related Works}\label{sect:RelWorks}
The theoretical study of model counting dates back to at least 1979~\cite{V79}, where the problem has been shown to be $\# \mathsf{P}$-complete;
meanwhile, 
the first version of $\mathsf{ApproxMC}$ has been proposed in 2013~\cite{ChakrabortyMV13}. 
A summary of historical cornerstones in the theory of model counting and its hashing-based approximate solvers can be found in e.g.,~\cite{YangM23,ChakrabortyMV16}. 

$\mathsf{ApproxMC}$ has undergone continuous updates since its first release~\cite{ChakrabortyMV13}. 
The subroutine $\mathsf{LogSATSearch}$ in $\mathsf{ApproxMCCore}$ is due to the second update~\cite{ChakrabortyMV16}; 
The third and fourth version proposed an efficient handling of CNF formulas combined with XOR constraints~\cite{SM19,SoosGM20}; 
then sparse hash families for $\mathsf{ApproxMC}$ in the fifth version~\cite{AM20}, and 
the rounding method in the sixth version~\cite{YangM23}, have been added respectively. 
While significant performance improvements have been made through these updates, 
there has been a limited focus on systematically understanding our analysis tool for the key component of $\mathsf{ApproxMC}$, that is, the bounding argument on  $\mathsf{ApproxMCCore}$. Our paper is intended to fill this gap, and serve as a theoretical basis of future developments. 

Yet another new variant of $\mathsf{ApproxMC}$, called $\mathsf{ApproxMC7}$, has been released rather recently~\cite{PoteM025}. The experiments in the paper demonstrate better runtime performance of $\mathsf{ApproxMC7}$ over $\mathsf{ApproxMC6}$; however, $\mathsf{ApproxMC7}$ requires the assumption $\varepsilon > 1$ for its correctness. 
Extending our formalism to include $\mathsf{ApproxMC7}$ as a particular instance is left for future work. 
Indeed, $\mathsf{ApproxMC7}$ resembles $\mathsf{ApproxMC6}$ except for the different choice of internal parameters---for example, its core counter %
can be viewed as $\mathsf{ApproxMC6Core}$ with specific values of $\thresh$ and rounding parameters. However, it sets $\thresh = 1$, which falls outside the range of our analysis; there are also subtle differences in the derivation of the error probability bounds. Because of these, 
our current formalism cannot instantiate $\mathsf{ApproxMC7}$.